\documentclass[lettersize,journal]{IEEEtran}

\usepackage{booktabs}
\usepackage{graphicx} 

\IEEEoverridecommandlockouts
\usepackage{cite}
\usepackage{amsmath,amssymb,amsfonts}
\usepackage{graphicx}
\usepackage{amsthm}
\usepackage{textcomp}
\usepackage{xcolor}
\def\BibTeX{{\rm B\kern-.05em{\sc i\kern-.025em b}\kern-.08em
		T\kern-.1667em\lower.7ex\hbox{E}\kern-.125emX}}
\usepackage{bm}
\usepackage{caption}
\usepackage{textcomp}
\usepackage{xcolor}
\usepackage{color}
\usepackage[normalem]{ulem}
\usepackage{graphicx,cite,amsmath,amssymb,hhline,booktabs,stfloats,bm}
\usepackage{amssymb}
\usepackage{verbatim}
\usepackage{subfigure}
\usepackage[linesnumbered,ruled,lined]{algorithm2e}
\usepackage{tabu}
\usepackage{amsthm}
\usepackage{caption}
\usepackage{lipsum}
\usepackage{subfigure}

\usepackage{stfloats}
\usepackage{booktabs}



\def\BibTeX{{\rm B\kern-.05em{\sc i\kern-.025em b}\kern-.08em
		T\kern-.1667em\lower.7ex\hbox{E}\kern-.125emX}}

\begin{document}

\title{Secure and Energy-Efficient Wireless Agentic AI  Networks \\
 

\thanks{{Yuanyan Song, Kezhi Wang, and Xinmian Xu are with the Department of Computer Science, Brunel University of London, UB8 3PH, UK. Xinmian Xu is also with Nanjing University of Posts and Telecommunications.  
        } 
}}
\author{ \IEEEauthorblockN{ Yuanyan Song, Kezhi Wang,  \IEEEmembership{Senior Member, IEEE}, Xinmian Xu }
\IEEEauthorblockN{  }

}
\maketitle
\markboth{IEEE Transactions on Mobile Computing,~Vol.~XX, No.~X, XX~XXXX}%
{  \MakeLowercase{\textit{et al.}}}

	\begin{abstract}
    In this paper, we introduce a secure wireless agentic AI network comprising one supervisor AI agent and multiple other AI agents to provision quality of service (QoS) for users' reasoning tasks while ensuring confidentiality of private knowledge and reasoning outcomes. Specifically, the supervisor AI agent can dynamically assign other AI agents to participate in cooperative reasoning, while the unselected AI agents act as friendly jammers to degrade the eavesdropper's interception performance. To extend the service duration of AI agents, an energy minimization problem is formulated that jointly optimizes AI agent selection, base station (BS) beamforming, and AI agent transmission power, subject to latency and reasoning accuracy constraints. To address the formulated problem, we propose two resource allocation schemes, ASC and LAW, which first decompose it into three sub-problems. Specifically, ASC optimizes each sub-problem iteratively using the proposed alternating direction method of multipliers (ADMM)-based algorithm, semi-definite relaxation (SDR), and successive convex approximation (SCA), while LAW tackles each sub-problem using the proposed large language model (LLM) optimizer within an agentic workflow. The experimental results show that the proposed solutions can reduce network energy consumption by up to 59.1\% compared to other benchmark schemes. Furthermore, the proposed schemes are validated using a practical agentic AI system based on Qwen, demonstrating satisfactory reasoning accuracy across various public benchmarks. 

	\end{abstract}
	

\section{Introduction}
\IEEEPARstart{R}{ecently}, AI agents and the emerging paradigm of agentic AI, empowered by the remarkable capabilities of large language models (LLMs), have gained significant attention \cite{refnew1}. Leveraging the technical advantages of LLMs in complex reasoning, multi-step planning, and in-context learning, AI agents have evolved from passive responders to pre-defined tasks into autonomous entities capable of active environmental perception and automated tool invocation \cite{ref1}, \cite{ref2}. This technological leap significantly improves the automation of AI systems, enabling AI agents to dynamically execute specific tasks by interacting with dynamic environments and utilizing appropriate AI tools with minimal human intervention \cite{ref17}. To further address highly complex objectives, agentic AI frameworks have been proposed to dynamically coordinate multiple AI agents, forming a collaborative network rather than relying on a monolithic agent \cite{ref7}. Such architectures can substantially improve quality of service (QoS) and robustness for intricate tasks, as they allow the aggregation of diverse skill sets and the verification of outputs from different AI agents through cooperative mechanisms \cite{refnew2}.

\begin{figure}[!h]
		\centering
		\includegraphics[width=\linewidth]{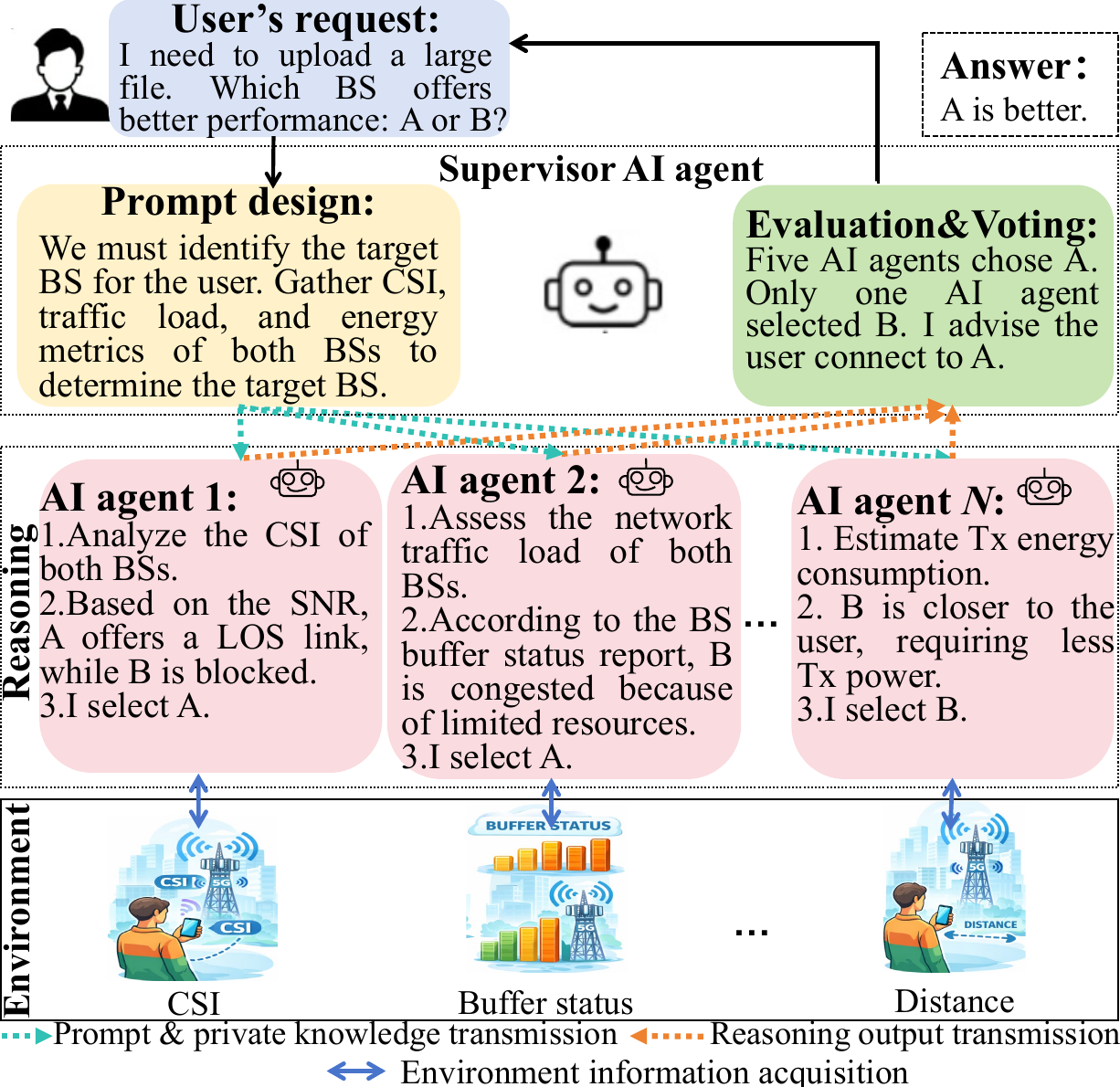}
		\caption{A typical workflow of agentic AI to execute a specific reasoning task.}
		\label{fig:typical workflow}
\end{figure}

Agentic AI is increasingly recognized as a promising architecture for QoS provisioning for the emerging reasoning tasks \cite{refnew6}. Unlike traditional services, these reasoning tasks are characterized by strong context-dependency, often requiring up-to-date external information from dynamic environments and private knowledge from users, making them difficult for conventional wireless networks to handle \cite{refnew3}. Agentic AI addresses these challenges by orchestrating multiple AI agents following dynamic workflows. A typical workflow is shown in Fig. \ref{fig:typical workflow}. The agentic AI network first interprets the user's request and subsequently dispatches it to multiple specialized AI agents. Each AI agent independently collects environmental data to derive its own reasoning conclusion. Finally, the system aggregates these diverse reasoning outputs to obtain a final result \cite{refnew4}, \cite{refnew5}. This cooperative workflow offers a potential pathway for mitigating the inaccuracies and incompleteness caused by single-agent hallucinations or limited data \cite{refnew7}. 

Thanks to the advancements in model lightweighting technologies, such as quantization and pruning, along with efficient post-training methods, the computational overhead of LLMs can be significantly reduced with negligible accuracy loss \cite{ref11}. These breakthroughs enable the deployment of LLMs on mobile devices (MDs) while maintaining satisfactory inference accuracy \cite{ref13} and the realization of wireless agentic AI networks. This promising paradigm can serve as a complement to traditional cloud-based agentic AI systems. By leveraging edge resources and the QoS provisioning capabilities of agentic AI, wireless agentic AI networks are expected to provide on-demand reasoning services with low latency, high accuracy and reliability for wireless networks. However, the implementation of wireless agentic AI networks is facing several critical technical challenges. First, MDs are typically constrained by limited battery capacities. How to dynamically coordinate network communication and computation resources to minimize the network's energy consumption while satisfying QoS constraints, such as reasoning accuracy and latency thresholds, remains a pressing problem. Second, the open nature of wireless channels renders the inter-AI agent communication links vulnerable to eavesdropping. This exposure risks the leakage of sensitive user information during the collaborative workflow, making the realization of confidential communication a pivotal research focus in agentic AI networks.

Motivated by the aforementioned challenges, this paper investigates secure communications for wireless agentic AI networks and optimizes communication and computation resources scheduling for energy minimization and reasoning tasks QoS provisioning. The main contributions of this work are summarized as follows:
	\begin{itemize}
		\item[1)] This article proposes a novel secure wireless agentic AI network, where one supervisor AI agent can dynamically select other AI agents to participate in cooperative reasoning, while other unselected AI agents perform as friendly jammers to ensure the confidentiality of private knowledge and reasoning outcomes. The energy minimization problem is formulated by jointly considering AI agent selection, BS beamforming, AI agent transmission powers and QoS constraints. The formulated problem is extremely challenging to tackle due to the coupling of integer variables and non-convex constraints.
        \item[2)] We first propose a resource allocation scheme called ASC to solve the formulated problem. In particular, the original problem is decoupled into three sub-problems, which are iteratively tackled by the proposed alternating direction method of multipliers (ADMM)-based algorithm, semi-definite relaxation (SDR) and successive convex approximation (SCA), respectively. As such, the feasible solution of the original problem can be obtained.
        \item[3)] Another LLM-based resource allocation scheme called LAW is proposed to decouple the original problem into three sub-problems. Each sub-problem can be solved by the proposed LLM optimizer following an agentic workflow. LAW is an LLM-based scheme designed to be directly implemented by the supervisor AI agent.
		\item[4)] Simulation results verify the effectiveness of ASC and LAW. The results demonstrate that the proposed resource allocation schemes significantly reduce the energy consumption of the proposed secure wireless agentic AI network. Furthermore, the impact of key system parameters, such as different QoS requirements and BS configurations, is investigated.
        \item[5)] We develop a practical agentic AI system based on the Qwen LLM series. Through extensive experiments on multiple public benchmarks, we validate that the proposed resource allocation schemes can achieve satisfactory reasoning accuracy.
	\end{itemize}

The rest of this article is organized as follows. The related works are demonstrated in Section \ref{related works}. The proposed secure wireless agentic AI network and formulated energy minimization problem are demonstrated in Section \ref{system model}. The proposed resource allocation schemes are introduced in \ref{proposed solution}. The performance evaluation regarding the proposed solutions along with numerous selected benchmark schemes is shown in Section \ref{performance evaluation}, followed by Section \ref{Conclusion} concluding this article.

\section{Related Works}\label{related works}
In this section, we report some related works regarding MD-deployed LLMs, agentic AI-empowered reasoning and some open challenges of wireless agentic AI networks.

\subsection{MD-deployed LLMs}
The deployment of LLMs in wireless networks is hindered by the disparity between the colossal scale of these models and the limited computation and memory resources of MDs. Some state-of-the-art models like GPT-4 contain approximately 1.8 trillion parameters, with next-generation LLMs such as GPT-5 expected to be higher \cite{ref9}. To overcome this issue, model quantization and pruning have been recognized as effective methods to reduce the high dimensionality of LLMs. \cite{ref11} and \cite{ref10} introduced advanced quantization methods, utilizing approximate second-order information and activation smoothing techniques, to significantly reduce model widths and parameter quantities while maintaining satisfactory reasoning capabilities. The authors of \cite{ref12} proposed an efficient fine-tuning and pruning strategy that effectively reduces the active dimensions and memory usage of LLMs, enabling a 65B parameter model to run on resource-limited hardware with negligible performance degradation. The practical viability of these optimizations was further investigated in \cite{ref13}, which conducted extensive experiments across various MDs, verifying that such lightweighted LLMs can be potentially deployed in resource-constrained wireless networks.

Post-training is of great importance to promise the performance of MD-deployed LLMs and enhance their accuracy when executing specific reasoning tasks. Scaling Law stands as a fundamental theoretical framework for predicting LLM training performance. The authors of \cite{refnew8} proposed empirical scaling laws for LLMs performance on the cross-entropy loss, which is one of the most important metrics of LLMs training. They pointed out that the loss scales follow a power-law with model size, dataset size, and the amount of computation resources
used for training. The authors of \cite{refnew9} investigated the scaling behavior of LLMs in different learning settings. They reported that downstream performance follows a predictable log-law and providing sufficient distributional alignment between the pretraining and downstream datasets is important. The authors of \cite{refnew10} focused on scaling law of reasoning capability enhancement in post-training and reported that lower-dimensional LLMs can achieve better reasoning accuracy by increasing the quantity of high-quality training data. The authors of \cite{refnew11} studied the scaling behaviors of LLMs during deep reinforcement learning (DRL)-based post-training. Through extensive experiments across the Qwen2.5 models ranging from 0.5B to 72B parameters, they proposed a scaling law for mathematical reasoning and validated its accuracy. Scaling law provides essential guidance for MD-deployed LLM systems formulation. How to realize a satisfactory reasoning accuracy while enhancing computational and energy efficiency needs further investigation.

\subsection{Agentic AI-empowered reasoning}
Reasoning tasks constitute a significant proportion of user requests, requiring complex logical deduction rather than simple data computation. Agentic AI has been recognized as a promising framework to provision QoS of these tasks. The authors of \cite{refnew12} proposed a novel “chain-of-thought” framework. Experiments verified that the proposed framework could significantly enhance the accuracy of single AI agent when executing reasoning tasks. Based on this idea, the authors of \cite{refnew4} proposed the ReAct framework, which enables AI agents to invoke external tools to gather real-time environment information. Experiments demonstrated that the proposed framework can effectively reduce hallucinations and further improve the reliability of the reasoning process. The authors of \cite{refnew13} proposed a novel self-consistency strategy to replace the greedy decoding method employed in “chain-of-thought” framework. By sampling diverse reasoning paths, this method effectively exploits the redundancy in reasoning processes to further enhance reasoning accuracy. The authors of \cite{refnew14} proposed a novel agentic AI-based reasoning framework and pointed out that the collaboration among multiple AI agents during the reasoning process can significantly enhance the accuracy and reliability of reasoning tasks. 

Majority voting is a widely adopted aggregation scheme in multi-agent reasoning, where the system selects the final answer based on the highest support among independent AI agents. The authors of \cite{refnew15} analyzed the multi-agent debate process and revealed that majority voting contributes to the primary performance gains. Similarly, the authors of \cite{refnew16} compared various decision protocols and demonstrated that voting schemes tend to yield larger improvements for reasoning tasks. As such, majority voting is regarded as an effective scheme to achieve satisfactory reasoning accuracy without significantly increasing network complexity.

\subsection{Open challenges of wireless agentic AI networks}
Resource allocation schemes are important for QoS provisioning in wireless agentic AI networks. Since agentic AI requires continuous data exchange between AI agents and intensive model inference, joint optimization of communication and computing resources becomes essential to satisfy latency and accuracy requirements of complex reasoning tasks. The authors of \cite{refnew17} proposed a DRL-based resource allocation scheme for edge-deployed LLM inference. Experiments verified that the proposed scheme could reduce inference latency while realizing satisfactory execution success rate. The authors of \cite{ref22} formulated a cooperative LLM inference model. The throughput maximization problem considering batch scheduling and resource allocation was formulated and an efficient heuristic algorithm was proposed to tackle the formulated problem. The simulation results verified that the proposed algorithm can reduce time complexity significantly. The authors of \cite{refnew18} proposed a wireless distributed mixture of experts LLM network, and formulated inference latency minimization problem subject to inference accuracy constraint. A dynamic expert selection policy with bandwidth allocation was proposed. Experiments verified that the proposed scheme can reduce the inference latency without compromising LLM performance. Although some recent works have investigated resource allocation for LLM inference in wireless networks, existing schemes cannot satisfy the unique requirements of wireless agentic AI networks. First, the collaborative nature of agentic AI involves significantly more complex data interaction between AI agents compared to standard LLM inference. Second, current approaches often overlook the energy optimization of AI agents. How to minimize network energy consumption while maintaining reasoning accuracy remains a critical unresolved issue. Furthermore, security issue is a major concern in wireless agentic AI networks. The frequent data exchange among AI agents over open wireless channels increases vulnerability to eavesdropping. How to protect the confidentiality of sensitive user data and reasoning results during inter-agent communication remains a significant challenge.

\section{System Model and Problem Formulation}\label{system model}
\begin{figure*}[!t]
    \centering
    \includegraphics[width=0.82\textwidth]{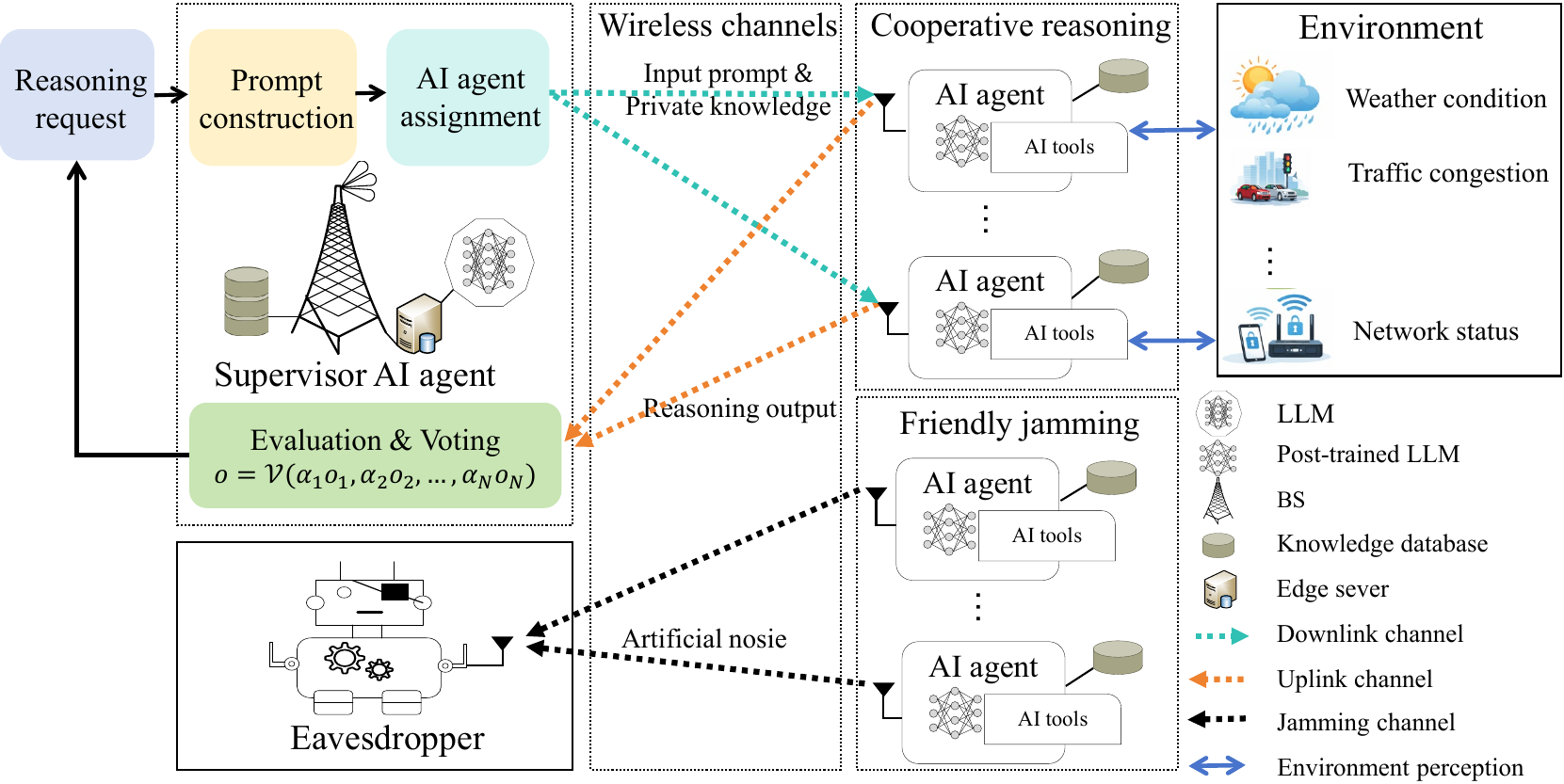}
    \caption{The framework of a secure wireless agentic AI network.}
    \label{fig:network}
\end{figure*}

As shown in Fig. \ref{fig:network}, we propose a secure wireless agentic AI network, where a supervisor AI agent $A_s$ and a set of $\mathcal{N}=\{A_1,A_2,\cdots,A_N\}$ AI agents dynamically form an agentic AI network for cooperative reasoning. $A_s$ is deployed on an edge server integrated with an $L$-antenna BS. Each $A_n$ runs on its respective single-antenna MD and can communicate with $A_s$ via wireless channels. $A_n$ is equipped with a post-trained LLM and AI tools, which can be utilized for specific reasoning tasks and environment perception, respectively. The knowledge database is hosted by $A_s$ to provide up-to-date private knowledge for $A_n$'s retrieval-augmented reasoning. Moreover, there is a single-antenna eavesdropper (eve) in the network attempting to obtain the reasoning task information, private knowledge, and reasoning outputs. 

Consider a reasoning task $U \triangleq \{s,D,o\}$, where $s$ is the length of input prompt, $D$ represents the data size of required private knowledge and $o$ denotes the reasoning output.  $A_s$ dispatches other AI agents to execute the reasoning task. Let $\alpha_n$ be the AI agent selection indicator, where $\alpha_n=1$ represents $A_s$ selecting $A_n$ to execute $U$ while $\alpha_n=0$ otherwise. $A_s$ transmits the input prompt and required private knowledge to selected AI agents simultaneously. Each selected AI agent can start task execution after receiving input prompt and private knowledge and returns  the reasoning output to $A_s$. For unselected AI agents, they transmit artificial noise to function as friendly jammers in the downlink and uplink transmission processes. 

\subsection{Secure communication model}\label{convert model}
$A_s$ transmits the input prompt and private knowledge to selected AI agents simultaneously, while other unselected AI agents function as friendly jammers. Assume that $A_n$ is selected to execute reasoning task and $A_{n'}$ injects artificial noise to degrade eve’s interception performance. Let $\bm{h}_\textit{BS, n} \in \mathbb{C}^{L\times 1}$, $\bm{h}_\textit{BS, eve} \in \mathbb{C}^{L\times 1}$, ${h}_\textit{n', n}$ and ${h}_\textit{n', eve}$ be the channel gain between BS and $A_n$, BS and eve, $A_n$ and $A_{n'}$, $A_{n'}$ and eve, respectively\footnote{In practical wireless scenarios, the eve typically remains passive and traditional channel state information (CSI) acquisition methods are ineffective for estimating the link between the transmitter and eve. Recently, integrated sensing and communications (ISAC) has been recognized as a potential framework to overcome this issue \cite{refnew20}, \cite{refnew21}. By employing pre-sensing stages, including beam scanning or environment probing, the transmitter can estimate eve’s angular and geometric parameters. This allows for the precise estimation, or a reliable approximation, of the eve's CSI even without its cooperation. Following this rationale, we assume that the CSI between the BS and eve, each AI agent and eve, is available for the design of transmission strategies.}. Denote the downlink beamforming vector by $\bm{w}$. One has
\begin{equation}
    \begin{aligned}
        ||\bm{w}||^2 \le p_\textit{BS}^{tr},
    \end{aligned}
    \label{eq:beamforming constraint}
\end{equation}
where $p_\textit{BS}^{tr}$ is the transmission power of BS. The received signal at $A_n$ can be given as
\begin{equation}
    \begin{aligned}
        y_n=\bm{h}_\textit{BS,n}^H \bm{w} x_\textit{BS}+\sum_{n'\in \mathcal{N}}(1-\alpha_{n'}){h}_\textit{n', n} \sqrt{p_{n'}^{tr}} v_{n'}+\sigma, n \in \mathcal{N},
    \end{aligned}
\end{equation}
where $x_\textit{BS}$ represents the transmitted data symbol of BS with $\mathbb{E}\{|x_\textit{BS}|^2\}=1$. $v_{n'}$ is the artificial noise with $v_n \sim \mathcal{CN}(0,1)$. $p_{n'}^{tr}$ denotes the transmission power of $A_{n'}$. $\sigma$ represents the noise variance. The downlink data rate at $A_n$ can be given as
\begin{equation}
    \begin{aligned}
        r_n^{dl}={\rm log}_2(1+\dfrac{||\bm{h}_\textit{BS, n}^H \bm{w}||^2}{\sum_{n'\in \mathcal{N}}(1-\alpha_{n'}){h}_\textit{n', n}^2 {p_{n'}^{tr}}+\sigma^2}), n \in \mathcal{N},
    \end{aligned}
    \label{eq:An downlink data rate}
\end{equation}
The achievable downlink data rate can be given as
\begin{equation}
    \begin{aligned}
        r^{dl}=\underset{\substack{\forall n \in \mathcal{N} \\ \alpha_n=1}}{\min}\{r_n^{dl}\}.
    \end{aligned}
    \label{eq:achievable rate}
\end{equation}
The downlink eavesdropping data rate at eve can be given as
\begin{equation}
    \begin{aligned}
        r_\textit{eve}^{dl}={\rm log}_2(1+\dfrac{||\bm{h}_\textit{BS, eve}^H \bm{w}||^2}{\sum_{n'\in \mathcal{N}}(1-\alpha_{n'}){h}_\textit{n', eve}^2 {p_{n'}^{tr}}+\sigma^2}). 
    \end{aligned}
    \label{eq:eve downlink data rate}
\end{equation}
Based on \eqref{eq:achievable rate} and \eqref{eq:eve downlink data rate}, the downlink secrecy capacity is
\begin{equation}
    \begin{aligned}
        c^{dl}=[r^{dl}-r_\textit{eve}^{dl}]^+, 
    \end{aligned}
    \label{eq:downlink secrecy capacity}
\end{equation}
where $[\cdot]^+=\max\{\cdot,0\}$. Let $B$ be the bandwidth resource of BS. The latency for BS to transmit the input prompt and private knowledge can be given as
\begin{equation}
    \begin{aligned}
        t^{dl}=\dfrac{s+D}{B \cdot c^{dl}}.
    \end{aligned}
    \label{eq:downlink latency}
\end{equation}

After all selected AI agents finish task execution, they can transmit their reasoning outputs back to $A_s$ simultaneously. Let $x_n$ be the transmitted data symbol of $A_n$ with $\mathbb{E}\{|x_n|^2\}=1$. The received signal at BS is
\begin{equation}
\resizebox{\linewidth}{!}{ 
    $\begin{aligned}
        \bm{y}_\textit{BS}=\sum_{n \in \mathcal{N}}\alpha_n \bm{h}_\textit{BS, n}^H \sqrt{p_n^{tr}} x_n+\sum_{n' \in \mathcal{N}}(1-\alpha_{n'})\bm{h}_\textit{BS, n'}^H\sqrt{p_{n'}^{tr}} v_{n'}+\sigma\bm{I},
    \end{aligned}$
}
\label{eq:recieved data}
\end{equation}
where $p_n^{tr}$ represents the transmission power of $A_n$. $\bm{I}\in \mathbb{C}^{L\times1}$ is a unit vector. The uplink data rate of BS to decode reasoning output from $A_n$ is
\begin{equation}
    \begin{aligned}
        r_n^{ul}={\rm log}_2(1+\dfrac{||\bm{h}_\textit{BS,n}^H \sqrt{p_n^{tr}}||^2}{\sum_{n' \in \mathcal{N} \setminus \{n\}}||\bm{h}_\textit{BS,n'}^H\sqrt{p_{n'}^{tr}}||^2+\sigma^2}), n \in \mathcal{N}.
    \end{aligned}
    \label{eq:uplink data rate}
\end{equation}
 The uplink eavesdropping data rate at eve can be given as
 \begin{equation}
     \begin{aligned}
         r_\textit{n, eve}^{ul}={\rm log}_2(1+\dfrac{|{h}_\textit{n,\;eve}|^2 p_n^{tr}}{\sum_{n' \in \mathcal{N}\setminus \{n\}}{h}_\textit{n',\;eve}^2{p_{n'}^{tr}}+\sigma^2}), n \in \mathcal{N}.
     \end{aligned}
     \label{eq:uplink eve data rate}
 \end{equation}
 Based on \eqref{eq:uplink data rate} and \eqref{eq:uplink eve data rate}, the uplink secrecy capacity is
 \begin{equation}
     \begin{aligned}
         c_n^{ul}=[r_n^{ul}-r_\textit{n,eve}^{ul}]^+. 
     \end{aligned}
 \end{equation}
As such, the latency for $A_n$ to transmit reasoning output is
\begin{equation}
    \begin{aligned}
        t_n^{ul}=\dfrac{o_n}{B \cdot c_n^{ul}}, n \in \mathcal{N},
    \end{aligned}
\end{equation}
where $o_n$ represents the reasoning output of $A_n$.

\subsection{Reasoning model}
Consider the cooperative reasoning process. The number of selected AI agents cannot be less than the predetermined threshold $N_{min}$. One has
\begin{equation}
    \begin{aligned}
        \sum_{n \in \mathcal{N}} \alpha_n \ge N_{min}.
    \end{aligned}
    \label{eq:number constraint}
\end{equation}

Let $k_n$ be the dimension of the LLM equipped by $A_n$. To enhance the reasoning capabilities of specific tasks, AI agent equipped LLMs need to undergo post-training. According to the scaling law of LLM training, the accuracy of an AI agent in completing one specific type of reasoning task can be predicted by its LLM model scale and the computational resources allocated during post-training \cite{refnew8}, \cite{refnew9}, \cite{refnew10}, which is
\begin{equation}\label{eq: accuracy}
    \begin{aligned}
        \textit{Acc}_n=1-{\rm exp}(- M(k_n) \cdot \log C_n + R(k_n)), n \in \mathcal{N}.
    \end{aligned}
\end{equation}
In \eqref{eq: accuracy}, $M(k_n)$ represents the learning efficiency, which is a function of $k_n$. $C_n$ represents the training budget, including training tokens and computation resources, used during training of $A_n$. $R(k_n)$ is a constant related to scale and structure of the LLM equipped by $A_n$.


To guarantee the QoS of cooperative reasoning, the average accuracy of selected AI agents cannot be less than a predefined threshold $\textit{Acc}_{min}$. One has
\begin{equation}
    \begin{aligned}
        \dfrac{\sum_{n \in \mathcal{N}} \alpha_n\cdot \textit{Acc}_{n}}{\sum_{n \in \mathcal{N}} \alpha_n} \ge \textit{Acc}_{min}.
    \end{aligned}
    \label{eq:loss constraint}
\end{equation}

In the cooperative reasoning, $A_n$ first employs AI tools to collect environmental data, leveraging up-to-date sensory data to ensure the accuracy of its reasoning. The tool calling information is generated by its LLM after receiving the input prompt and private knowledge. After obtaining required environmental contexts, $A_n$ can execute the assigned reasoning task. The inference latency of $A_n$ is
	\begin{equation}
		\begin{aligned}
		t_{n}^\textit{inference}=\frac{(\eta+1)k_n+k_n (q_n+o_n)+k_n(q_n^2+o_n^2)}{f_n},n\in \mathcal{N},
		\end{aligned}
		\label{eq:t_calling}
	\end{equation}
where $q_n$ denotes the length of the tool calling information. $\eta$ represents the floating point operations (FLOPs) required by the network activation process of LLM \cite{ref22}. $f_n$ represents computing capability.

Then, $A_n$ transmits reasoning output to $A_s$ for evaluation. After receiving reasoning outputs of assigned AI agents, $A_s$ can obtain the final output using a voting function $\mathcal{V}(\cdot)$, which can be given as $o=\mathcal{V}(\alpha_1 o_1, \alpha_2 o_2,\cdots,\alpha_n o_n)$ \cite{refnew15}.

\subsection{The overall process}
 $A_s$ selects other AI agents to participate in cooperative reasoning and other unselected AI agents perform as friendly jammers. Then, $A_s$ transmits input prompt and private knowledge to selected AI agents simultaneously. After all selected AI agents finish executing the assigned reasoning task, they transmit reasoning output back to $A_s$ for evaluation simultaneously. Let $\tau$ be the maximum execution latency. One has the latency constraint
\begin{equation}
    \begin{aligned}
        t^{dl}+\underset{\substack{\forall n \in \mathcal{N} \\ \alpha_n=1}}{\max}\{t_n^\textit{inference}\}+\underset{\substack{\forall n \in \mathcal{N} \\ \alpha_n=1}}{\max}\{t_n^{ul}\} \le \tau.
    \end{aligned}
    \label{eq:latency constraint}
\end{equation}
The transmission energy consumption of $A_n$ is formulated as
\begin{equation}
\resizebox{\linewidth}{!}{ 
    $\begin{aligned}
        E_n^{tr}=p_n^{tr}(\alpha_n t_n^{ul}+(1-\alpha_n)(t^{dl}+\underset{\substack{\forall n \in \mathcal{N} \\ \alpha_n=1}}{\max}\{t_n^{ul}\})), n\in \mathcal{N}.
    \end{aligned}$}
\label{eq:Etr}
\end{equation}
Let $E_n^\textit{exe}$ be the execution energy consumption of $A_n$, which can be given as
	\begin{equation}
		\begin{aligned}
			E_n^\textit{exe}=\alpha_n p_{n}^\textit{exe}t_{n}^\textit{inference},n \in \mathcal{N},
		\end{aligned}
		\label{eq:E_exe}
	\end{equation}
	where $p_{n}^\textit{exe}$ represents the computing power of $A_n$.
    
\subsection{Problem formulation}
In this paper, we aim to minimize the overall energy consumption of AI agents by jointly optimizing AI agent selection $\bm{\alpha}\triangleq\{\alpha_n, n \in \mathcal{N}\}$, BS beamforming vector $\bm{w}$, and AI agent transmission powers $\bm{p} \triangleq \{p_n^{tr}, n \in \mathcal{N}\}$, which can be formulated as
\begin{subequations} \label{prob}
		\begin{align}
			&\underset{\bm{\alpha},\bm{w},\bm{p}}{\min}\;\sum_{n \in \mathcal{N}}E_n^\textit{exe}+E_n^\textit{tr},\\
			&{\rm s.\;t.}\;\;\; \eqref{eq:beamforming constraint},\eqref{eq:number constraint},\eqref{eq:loss constraint},\eqref{eq:latency constraint}, \notag\\
			&\;\;\;\;\;\;\;\;\alpha_n\in\{0,1\}, n \in \mathcal{N},\label{constraint: AI agent binary}\\
            &\;\;\;\;\;\;\;\;0\le p_n^{tr}\le p_n^\textit{tr,max}, n \in \mathcal{N},\label{constraint: transmission power}
		\end{align}
	\end{subequations}
where \eqref{constraint: AI agent binary} is the binary constraint of AI agent selection indicator $\bm{\alpha}$ and \eqref{constraint: transmission power} ensures the transmission power of each $A_n$ cannot exceed  $p_n^\textit{tr,max}$.

\section{ The proposed solutions} \label{proposed solution}
The formulated optimization problem is a  mixed-integer non-linear optimization problem, which is extremely difficult to tackle. To efficiently solve this problem, this section introduces two resource allocation scheme, i.e., ASC, a \textbf{A}DMM-\textbf{S}DR-SCA \textbf{C}oordinated sheme and LAW, a \textbf{L}LM-enabled \textbf{A}gentic \textbf{W}orkflow-based scheme. The detailed information regarding the proposed solutions is given as follows.
\subsection{ASC}
In this subsection, we first introduce ASC to solve \eqref{prob}. The original optimization problem is decoupled into three sub-problems, i.e., AI agent selection, beamforming design and transmission power optimization sub-problems. In particular, the AI agent selection sub-problem is solved by the proposed ADMM-based algorithm, while the beamforming design and transmission power optimization sub-problems are tackled by SDR and SCA, respectively. Each sub-problem is solved in an iterative manner until convergence.
\subsubsection{AI agent Selection}
For any given feasible $\bm{w}$ and $\bm{p}$, \eqref{prob} can be reduced as
\begin{equation} \label{subprob1}
		\begin{aligned}
			&\underset{\bm{\alpha}}{\min}\;\sum_{n \in \mathcal{N}}E_n^\textit{exe}+E_n^\textit{tr},\\
			&{\rm s.\;t.}\;\;\; \eqref{eq:number constraint},\eqref{eq:loss constraint},\eqref{eq:latency constraint},\\
			&\;\;\;\;\;\;\;\;\alpha_n\in\{0,1\}, n \in \mathcal{N}.
		\end{aligned}
\end{equation}

Problem \eqref{subprob1} is extremely challenging to tackle due to the complex dependency of the objective function. First, $\bm{\alpha}$ is closely coupled within $E_n^{tr}$ and in constraints \eqref{eq:number constraint} and \eqref{eq:loss constraint}. The term $r^{dl}=\underset{\forall n \in \mathcal{N},\alpha_n=1}{\min}\{{\rm log}_2(1+\frac{||\bm{h}_\textit{BS, n}^H \bm{w}||^2}{\sum_{n'\in \mathcal{N}}(1-\alpha_{n'}){h}_\textit{n', n}^2 {p_{n'}^{tr}}+\sigma^2})\}$ is non-smooth, which is difficult to transformed into a closed-form algebraic function of $\bm{\alpha}$. Second, even with fixed $r^{dl}$, the binary nature of $\alpha_n \in \{0,1\}$ classifies \eqref{subprob1} as a combinatorial non-linear integer programming problem, which is known to be NP-hard. Widely used approaches, such as branch-and-bound or genetic algorithms may become inefficient to solve \eqref{subprob1}. To address these challenges, an ADMM-based algorithm is proposed. First, an auxiliary variable $\hat{\bm{\alpha}}\triangleq\{\hat{\alpha}_n, n \in \mathcal{N}\}$ is introduced, which should satisfy
\begin{equation}
    \begin{aligned}
        \hat{\alpha}_n={\alpha}_n, n\in \mathcal{N}.
    \end{aligned}
    \label{eq:duplicated constraint}
\end{equation}
One can transform $E_n^{tr}$ into $\hat{E}_n^{tr}$ via substituting $\bm{\alpha}$ in $r^{dl}$, $r_\textit{eve}^\textit{dl}$ and $\underset{\forall n \in \mathcal{N}, \alpha_n=1}{\max}\{\frac{o_n}{B\cdot [r_n^{ul}-r_\textit{n,eve}^{ul}]^+}\}$ by $\hat{\bm{\alpha}}$. Similarly, \eqref{eq:latency constraint} can be transformed as
\begin{equation}
    \begin{aligned}
         \hat{t}^{dl}+\underset{\substack{\forall n \in \mathcal{N} \\ \hat{\alpha}_n=1}}{\max}\{t_n^\textit{inference}\}+\underset{\substack{\forall n \in \mathcal{N} \\ \hat{\alpha}_n=1}}{\max}\{t_n^{ul}\} \le \tau,
    \end{aligned}
    \label{transformed latency constraint}
\end{equation}
where $\hat{t}^{dl}$ can be obtained via substituting $\bm{\alpha}$ by $\hat{\bm{\alpha}}$. As such, \eqref{subprob1} can be transformed into
\begin{equation} \label{subprob1.1}
		\begin{aligned}
			&\underset{\bm{\alpha},\hat{\bm{\alpha}}}{\min}\;\sum_{n \in \mathcal{N}}\hat{E}_n^\textit{exe}+E_n^\textit{tr},\\
			&{\rm s.\;t.}\;\;\; \eqref{eq:number constraint},\eqref{eq:loss constraint}, \eqref{eq:duplicated constraint},\eqref{transformed latency constraint}, \\
			&\;\;\;\;\;\;\;\;\alpha_n\in\{0,1\}, n \in \mathcal{N}.
		\end{aligned}
\end{equation}
The corresponding augmented Lagrangian function of \eqref{subprob1.1} can be formulated as
\begin{equation}
    \begin{aligned}
        \mathcal{L}=\sum_{n \in \mathcal{N}}(\hat{E}_n^\textit{exe}+E_n^\textit{tr})+\rho ||\bm{\alpha}-\hat{\bm{\alpha}}||_2^2,
    \end{aligned}
\end{equation}
where $\rho$ is the penalty coefficient of \eqref{eq:duplicated constraint}. One can observe that $\mathcal{L}$ is separable along
with $\bm{\alpha}$ and $\hat{\bm{\alpha}}$. \eqref{subprob1.1} can be solved by optimizing $\bm{\alpha}$ and $\hat{\bm{\alpha}}$ in an iterative manner.

(a). \textbf{The optimization of $\bm{\alpha}$}: Let $\hat{\bm{\alpha}}^{r_\textit{ADMM}-1}$ be the obtained solution of $\hat{\bm{\alpha}}$ in the $r_\textit{ADMM}-1$-th iteration. The optimization problem of $\bm{\alpha}$ in the the $r_\textit{ADMM}$-th iteration can be reduced as
\begin{equation}
    \begin{aligned}
        &\underset{\bm{\alpha}}{\min}\;\sum_{n \in \mathcal{N}}(\hat{E}_n^\textit{exe}+E_n^\textit{tr})+\rho ||\bm{\alpha}-\hat{\bm{\alpha}}^{r_\textit{ADMM}-1}||_2^2,\\
			&{\rm s.\;t.}\;\;\;\;\;\;\;\;\;\;\;\; \eqref{eq:number constraint},\eqref{eq:loss constraint}, \alpha_n\in\{0,1\}, n \in \mathcal{N}.
    \end{aligned}
    \label{subprob1.1.1}
\end{equation}
In this paper, a LR-based algorithm is proposed to tackle \eqref{subprob1.1.1}. Let $\mu$ and $\lambda$ be the Lagrangian multipliers of \eqref{eq:number constraint} and \eqref{eq:loss constraint}, respectively. The Lagrangian function of \eqref{subprob1.1.1} can be formulated as \eqref{Lagrangian function LR}.
\begin{figure*}[!t]
\normalsize
\begin{equation}
    \begin{split}
        \mathcal{L}^\textit{LR}(\bm{\alpha}, \mu, \lambda) &= \sum_{n \in \mathcal{N}}(\hat{E}_n^\textit{exe}+E_n^\textit{tr}) + \rho \|\bm{\alpha}-\hat{\bm{\alpha}}^{r_\textit{ADMM}-1}\|_2^2+ \mu (\sum_{n \in \mathcal{N}}\alpha_n-N_\textit{min}) + \lambda (\sum_{n \in \mathcal{N}}\alpha_n \textit{Acc}_n-\textit{Acc}_\textit{min}\sum_{n \in \mathcal{N}}\alpha_n)\\
        &=\sum_{n \in \mathcal{N}}  \rho \alpha_n^2+\alpha_n ( p_n^{tr}t_n^{ul}+p_n^\textit{exe}t_n^\textit{inference}-\frac{s+D}{B\cdot \hat{c}^{dl}}-\underset{\substack{\forall n \in \mathcal{N} \\ \hat{\alpha}_n^{r_\textit{ADMM}-1}=1}}{\max}\{\frac{o_n}{B\cdot c_n^{ul}}\}+\mu+\lambda(\textit{Acc}_n-\textit{Acc}_\textit{min})-2\rho\hat{\alpha}_n^{r_\textit{ADMM}-1} ) \\
        &\quad + \rho(\hat{\alpha}_n^{r_\textit{ADMM}-1})^2.
    \end{split}
    \label{Lagrangian function LR}
\end{equation}
\hrulefill
\vspace*{4pt}
\end{figure*}
In this way, \eqref{subprob1.1.1} can be reformulated as
\begin{equation}
    \begin{aligned}
        &\underset{\mu,\lambda}{\max}\;\;\;\;\{ \underset{\bm{\alpha}}{\min}\;\mathcal{L}^{LR}\},\\
			&{\rm s.\;t.}\;\alpha_n\in\{0,1\}, n \in \mathcal{N}, \mu \ge 0, \lambda \ge 0.
    \end{aligned}
    \label{transformed prob1.1.1}
\end{equation}
Let $\bm{\alpha}^{r_\textit{LR}-1}(\mu^{r_\textit{LR}-1}, \lambda^{r_\textit{LR}}) \triangleq \{\alpha_n^{r_\textit{LR}}, n \in \mathcal{N}\}$ be the $r_\textit{LR}$-th feasible solution to \eqref{transformed prob1.1.1}. For any given Lagrangian multipliers $\mu^{r_\textit{LR}-1}$ and $\lambda^{r_\textit{LR}-1}$, the update mechanism of $\bm{\alpha}^{r_\textit{LR}}(\mu^{r_\textit{LR}}, \lambda^{r_\textit{LR}})$ can be given as
\begin{equation}
\alpha_n^{{r_\textit{LR}}} =
    \begin{cases}
        1, & \rho+b_n^{{r_\textit{LR}-1}} \leq 0, \\
        0, & \text{otherwise},
    \end{cases}
    \label{update mechamism of alpha}
\end{equation}
where $b_n^{{r_\textit{LR}-1}}= p_n^{tr}t_n^{ul}+p_n^\textit{exe}t_n^\textit{inference}-\frac{s+D}{B\cdot \hat{c}^{dl}}-\underset{\substack{\forall n \in \mathcal{N} \\  \hat{\alpha}_n^{r_\textit{ADMM}-1}=1}}{\max}\{\frac{o_n}{B\cdot c_n^{ul}}\}+\mu^{{r_\textit{LR}-1}}+\lambda^{{r_\textit{LR}-1}}(\textit{Acc}_n-\textit{Acc}_\textit{min})-2\rho\hat{\alpha}_n^{r_\textit{ADMM}-1}$. One can utilize the subgradient method to update the Lagrangian multipliers in each iteration. In particular, the update mechanism of $\mu$ can be given as
\begin{equation}
    \begin{aligned}
        \mu^{r_\textit{LR}}=\max\{\mu^{r_\textit{LR}-1}+\psi^{r_\textit{LR}}(N_\textit{min}-\sum_{n \in \mathcal{N}}\alpha_n^{r_\textit{LR}}),0\},
    \end{aligned}
    \label{mu update mechanism}
\end{equation}
where $\psi^{r_\textit{LR}} \in (0,1)$ represents  the proportionality coefficient of $\mu^{r_\textit{LR}}$. The update 
mechanism of $\lambda$ can be given as
\begin{equation}
    \begin{aligned}
        \lambda^{r_\textit{LR}}=\max\{\lambda^{r_\textit{LR}-1}+\omega^{r_\textit{LR}}(\sum_{n \in \mathcal{N}}\alpha_n^{r_\textit{LR}}(\textit{Acc}_n-\textit{Acc}_\textit{min})),0\},
    \end{aligned}
    \label{lambda update mechanism}
\end{equation}
where $\omega^{r_\textit{LR}} \in (0,1)$ is the proportionality coefficient of $\lambda^{r_\textit{LR}}$. Let $r_\textit{LR}^\textit{max}$ be the maximum number of iterations of the proposed LR-based algorithm. The LR-based algorithm can be regarded as convergence when $r_\textit{LR}=r_\textit{LR}^\textit{max}$.

Due to the relaxation of constraints, the obtained $\bm{\alpha}^{r_\textit{LR}^\textit{max}}$ may not satisfy \eqref{eq:number constraint} and \eqref{eq:loss constraint}. To address this issue, a greedy method is employed to reconstruct $\bm{\alpha}^{r_\textit{LR}^\textit{max}}$. The modified AI agent selection indicator $\hat{\bm{\alpha}}^{r_\textit{LR}^\textit{max}}$ can be utilized to obtain the feasible solution $\bm{\alpha}^{r_\textit{ADMM}}$ of \eqref{subprob1.1.1}. 

(b). \textbf{The optimization of $\hat{\bm{\alpha}}$}: For any given $\bm{\alpha}^{r_\textit{ADMM}}$, \eqref{transformed prob1.1.1} can be reduced as
\begin{equation}
    \begin{aligned}
        &\underset{\hat{\bm{\alpha}}}{\min}\;\sum_{n \in \mathcal{N}}(\hat{E}_n^\textit{exe}+E_n^\textit{tr})+\rho ||\hat{\bm{\alpha}}-\bm{\alpha}^{r_\textit{ADMM}-1}||_2^2,\\
			&{\rm s.\;t.}\;\;\;\;\;\;\;\;\;\;\;\; \eqref{transformed latency constraint}, \hat{\alpha}_n\in\{0,1\}, n \in \mathcal{N}.
    \end{aligned}
    \label{subprob1.1.2}
\end{equation}
Problem \eqref{subprob1.1.2} is difficult to tackle. First, $\hat{r}^{dl}$ cannot be transformed into a close-form analytical expression of $\hat{\bm{\alpha}}$. Second, the binary constraint leads to traditional convex optimization algorithms hard to tackle \eqref{subprob1.1.2}. To settle this challenging optimization problem, we first relax $\hat{\bm{\alpha}}$ to the continuous domain. One has
\begin{equation}
    \begin{aligned}
        \hat{\alpha}_n \in [0,1], n \in \mathcal{N}.
    \end{aligned}
    \label{relaxed binary constraint}
\end{equation}
Then, a penalty-based method is utilized to transform $\hat{r}^{dl}$ into $\overline{r}^{dl}$. One has
\begin{equation}
    \begin{aligned}
        \overline{r}^{dl}\triangleq\underset{\forall n \in \mathcal{N}}{\min}\{\hat{r}_n^{dl}+(1-\hat{\alpha}_n) \Omega\},
    \end{aligned}
    \label{penality downlink rate}
\end{equation}
where $\hat{r}_n^{dl}$ can be obtained via substituting $\bm{\alpha}$ by $\hat{\bm{\alpha}}$. $\Omega$ is a sufficiently large penalty coefficient. This penalty term ensures that if $\hat{\alpha}_n = 0$, $r_n^{dl} +(1- \hat{\alpha}_n )\Omega$ becomes significantly larger than the data rates of AI agents, thereby effectively excluding $A_n$ from the minimization operation and $\hat{\alpha}_n = 1$ otherwise. After removing irrelevant parameters, \eqref{subprob1.1.2} can be rewritten as
\begin{equation}
    \begin{aligned}
        &\underset{\hat{\bm{\alpha}}}{\min}\sum_{n \in \mathcal{N}}(1-\alpha_n^{r_\textit{ADMM}}) ( \frac{s+D}{B\cdot [\overline{r}^{dl}-r_\textit{eve}^{dl}]^+}\\
        &\;\;\;\;\;+ \underset{\forall n \in \mathcal{N}}{\max}\{\frac{\hat{\alpha}_no_n}{B\cdot [r_n^{ul}-r_\textit{n,eve}^{ul}]^+}\} ) +\rho ||\hat{\bm{\alpha}}-\bm{\alpha}^{r_\textit{ADMM}-1}||_2^2,\\
			&{\rm s.\;t.}\;\;\;\;\;\;\;\;\;\;\;\;\;\;\;\;\;\;\;\;\;\; \eqref{transformed latency constraint}, \eqref{relaxed binary constraint}.
    \end{aligned}
    \label{transformed transformed subprob1.1.2}
\end{equation}
However, \eqref{transformed transformed subprob1.1.2} is still difficult to tackle due to the non-smooth objective function and non-convex constraint \eqref{transformed latency constraint}. An auxiliary variable $\hat{c}^{dl}$ is introduced. \eqref{transformed latency constraint} can be transformed into a convex constraint, which can be given as
\begin{equation}
    \begin{aligned}
         \frac{s+D}{\hat{c}^{dl}}+\underset{\forall n \in \mathcal{N}}{\max}\{\hat{\alpha}_nt_n^\textit{inference}\}+\underset{\forall n \in \mathcal{N}}{\max}\{\hat{\alpha}_nt_n^{ul}\} \le \tau.
    \end{aligned}
    \label{latecny constraint subprob1.1.2}
\end{equation}
As such, \eqref{transformed transformed subprob1.1.2} can be transformed into
 \begin{subequations} \label{transformed transformed transformed subprob1.1.2}
 \small
		\begin{align}
			&\underset{\hat{\bm{\alpha}},\hat{c}^{dl}}{\min}\;\sum_{n \in \mathcal{N}}(1-\alpha_n^{r_\textit{ADMM}}) ( \frac{s+D}{B\cdot \hat{c}^{dl}}+ \underset{\forall n \in \mathcal{N}}{\max}\{\frac{\hat{\alpha}_no_n}{B\cdot [r_n^{ul}-r_\textit{n,eve}^{ul}]^+}\} ) \notag \\
        &\;\;\;\;\;\;\;\;\;\;\;\;\;\;\; +\rho ||\hat{\bm{\alpha}}-\bm{\alpha}^{r_\textit{ADMM}-1}||_2^2,\\
			&{\rm s.\;t.}\;\;\;\;\;\;\;\;\;\;\;\;\;\;\;\;\;\;\;\;\;\; \eqref{relaxed binary constraint}, \eqref{latecny constraint subprob1.1.2}, \notag\\
            &\;\;\;\;\;\;\;\;\;\;\;\;\;\;\;\;\;\;\;\;\;\;\;\;\;\;\;\;\;\hat{c}^{dl} \ge 0,\label{introduced rate constraint1}\\
            &\;\;\;\;\;\;\;\;\;\;\;\;\;\;\;\;\;\;\;\;\;\;\;\;\;\;\;\;\;\hat{c}^{dl} \le \overline{r}^{dl}-r_\textit{eve}^{dl}.\label{introduced rate constraint2}
		\end{align}
\end{subequations}
However, problem \eqref{transformed transformed transformed subprob1.1.2} is still non-convex due to the non-convexity of \eqref{introduced rate constraint2}. To address this problem, SCA is utilized to construct the convex approximation of \eqref{introduced rate constraint2}. \eqref{introduced rate constraint2} can be rewritten as
\begin{equation}
      \begin{aligned}
         \hat{c}^{dl}+\hat{r}_\textit{eve}^{dl}+\underset{\forall n \in \mathcal{N}}{\max}\{-\hat{r}_n^{dl}-(1-\hat{\alpha}_n) \Omega\} \le 0.
      \end{aligned}
      \label{transformed introduced rate constraint2}
\end{equation}
One can observe that $-\hat{r}_n^{dl}$ is concave with respect to $\hat{\bm{\alpha}}$. The upper bound of $-\hat{r}_n^{dl}$ can be obtained via employing the first order Taylor expansion, which can be formulated as \eqref{lower bound of r_n}.
\begin{figure*}[!t]
    \begin{equation} \label{lower bound of r_n}
        \begin{aligned}
            \{-\hat{r}_n^{dl}\}^{ub}\triangleq & - {\rm log}_2\left(1+\dfrac{||\bm{h}_{\textit{BS}, n}^H \bm{w}||^2}{\sum_{n'\in \mathcal{N}}(1-\hat{\alpha}_{n'}^{r_{\textit{ADMM}}-1}){h}_{n', n}^2 {p_{n'}^{tr}}+\sigma^2}\right)\\
            & -\sum_{n'\in \mathcal{N}}\frac{||\bm{h}_{\textit{BS}, n}^H \bm{w}||^2{h}_{n', n}^2 {p_{n'}^{tr}} (\hat{\alpha}_{n'}-\hat{\alpha}_{n'}^{r_\textit{ADMM}-1})}{\left(\sum_{n'\in \mathcal{N}}(1-\hat{\alpha}_{n'}^{r_{\textit{ADMM}}-1}){h}_{n', n}^2 {p_{n'}^{tr}}+\sigma^2\right)\left(\sum_{n'\in \mathcal{N}}(1-\hat{\alpha}_{n'}^{r_{\textit{ADMM}}-1}){h}_{n', n}^2 {p_{n'}^{tr}}+\sigma^2+||\bm{h}_{\textit{BS}, n}^H \bm{w}||^2\right)}
        \end{aligned}
    \end{equation}
    \hrulefill
\end{figure*}

Accordingly, \eqref{transformed introduced rate constraint2} can be transformed into
\begin{equation}
    \begin{aligned}
        \hat{c}^{dl}+\hat{r}_\textit{eve}^{dl}+\underset{\forall n \in \mathcal{N}}{\max}\{\{-\hat{r}_n^{dl}\}^{ub}-(1-\hat{\alpha}_n) \Omega\}\le 0,
    \end{aligned}
    \label{convex transformed introduced rate constraint2}
\end{equation}
which is a convex constraint. Consequently, \eqref{transformed transformed transformed subprob1.1.2} can be transformed into
\begin{equation}\label{final subprob1.1.2}
    \begin{aligned}
        &\underset{\hat{\bm{\alpha}},\hat{c}^{dl}}{\min}\;\sum_{n \in \mathcal{N}}(1-\alpha_n^{r_\textit{ADMM}}) ( \frac{s+D}{B\cdot \hat{c}^{dl}}+ \underset{\forall n \in \mathcal{N}}{\max}\{\frac{\hat{\alpha}_no}{B\cdot [r_n^{ul}-r_\textit{n,eve}^{ul}]^+}\} )  \\
        &\;\;\;\;\;\;\;\;\;\;\;\;\;\;\; +\rho ||\hat{\bm{\alpha}}-\bm{\alpha}^{r_\textit{ADMM}-1}||_2^2,\\
			&{\rm s.\;t.}\;\;\;\;\;\;\;\;\;\;\;\;\;\;\;\;\;\;\;\;\;\; \eqref{relaxed binary constraint}, \eqref{latecny constraint subprob1.1.2}, \eqref{introduced rate constraint1}, \eqref{convex transformed introduced rate constraint2},
    \end{aligned}
\end{equation}
which is a convex optimization problem and can be efficiently solved. Let $\hat{\bm{\alpha}}^*\triangleq\{\hat{\alpha}_n^*, n \in \mathcal{N}\}$ be the obtained solution of \eqref{final subprob1.1.2}. We recover the binary values of $\hat{\bm{\alpha}}$ by setting $\hat{\alpha}_n^{r_\textit{ADMM}}=1$ if $\hat{\alpha}_n^* \ge 0.5$, and $\hat{\alpha}_n^{r_\textit{ADMM}}=0$ otherwise.

Let $r_\textit{ADMM}^\textit{max}$ be the maximum number of iterations of the proposed ADMM-based algorithm. The proposed ADMM-based algorithm can be regarded as convergence when $r_\textit{ADMM}=r_\textit{ADMM}^\textit{max}$. As such, the feasible AI agent selection $\bm{\alpha}^{r}$ can be obtained. The detailed information regarding the proposed ADMM-based algorithm can be found in \textbf{Algorithm \ref{ADMM}}.
\begin{algorithm}
		\caption{The framework of the proposed ADMM-based algorithm}
		\label{ADMM}
		Initialize $\bm{\alpha}^{0}$, $\hat{\bm{\alpha}}^{0}$, ${\lambda}^{0}$, ${\mu}^{0}$, $r_\textit{ADMM}=0$ and $r_\textit{LR}=0$.\\
        \While{$r_\textit{ADMM} \le r_\textit{ADMM}^\textit{max}$}{
		Fix $\hat{\bm{\alpha}}^{r_\textit{ADMM}}$, construct Lagrangian function according to \eqref{Lagrangian function LR}.\\
		\While{$r_\textit{LR} \le r_\textit{LR}^\textit{max}$}{
			Update $\bm{\alpha}^{r_\textit{LR}+1}$ according to \eqref{update mechamism of alpha}.\\
			Update ${\lambda}^{r_\textit{LR}+1}$ and ${\mu}^{r_\textit{LR}+1}$ according to \eqref{mu update mechanism} and \eqref{lambda update mechanism}.\\
			$r_\textit{LR} \leftarrow r_\textit{LR}+1$.\\
            }
            \If{$\bm{\alpha}^{r_\textit{LR}^\textit{max}}$ cannot satisfy \eqref{eq:number constraint} or \eqref{eq:loss constraint}}{Perform the greedy method to reconstruct $\bm{\alpha}^{r_\textit{LR}^\textit{max}}$.}
            $\bm{\alpha}^{r_\textit{ADMM}+1} \leftarrow \bm{\alpha}^{r_\textit{LR}^\textit{max}}$.\\
            Fix $\bm{\alpha}^{r_\textit{ADMM}}$, solve \eqref{final subprob1.1.2} and obtain $\hat{\bm{\alpha}}^*$.\\
            Obtain $\hat{\bm{\alpha}}^{r_\textit{ADMM}+1}$ by rounding $\hat{\bm{\alpha}}^*$ to the nearest integers.\\
            $r_\textit{ADMM} \leftarrow r_\textit{ADMM}+1$.\\
            }
		\textbf{Return} $\bm{\alpha}^{r_\textit{ADMM}^\textit{max}}$.\\
	\end{algorithm}

\subsubsection{Beamforming design}
For any given $\bm{\alpha}$ and $\bm{p}$ and according to \eqref{eq:downlink latency} and \eqref{eq:Etr}, \eqref{prob} can be reduced as
\begin{equation} \label{subprob3}
		\begin{aligned}
			&\underset{\bm{w}}{\max}\;\;c^{dl},\\
			&{\rm s.\;t.}\;\;\eqref{eq:beamforming constraint}.\\
		\end{aligned}
\end{equation}
$c^{dl}$ is non-convex with respect to $\bm{w}$ due to the existence of $\min\{\cdot\}$ function coupled in $r^{dl}$. To address this problem, an auxiliary variable, i.e., $\tilde{r}^{dl}$ is introduced linearize the objective function. \eqref{subprob3} can be transformed into
\begin{subequations} \label{transformed subprob3}
    \begin{align}
        &\underset{\bm{w},\tilde{r}^{dl}}{\max}\; \tilde{r}^{dl}-r_\textit{eve}^{dl},\\
        &{\rm s.\;t.}\;\;\;\;\;\; \eqref{eq:beamforming constraint},\notag\\
        &\;\;\;\;\;\;\;\tilde{r}^{dl} \le r_n^{dl}, n \in \mathcal{N}.\label{beamforming introduced constraint}
    \end{align}
\end{subequations}
In this paper, SDR is utilized to tackle \eqref{transformed subprob3}. One has $||\bm{h}_\textit{BS, n}^H \bm{w}||^2=\bm{h}_\textit{BS, n}^H \bm{w} \bm{w}^H \bm{h}_\textit{BS, n}={\rm tr}(\bm{h}_\textit{BS, n}^H \bm{w} \bm{w}^H \bm{h}_\textit{BS, n})={\rm tr}(\bm{h}_\textit{BS, n}^H \bm{h}_\textit{BS, n} \bm{w} \bm{w}^H)$. Let $\bm{H}_n=\bm{h}_\textit{BS, n}^H\bm{h}_\textit{BS, n} \in \mathbb{C}^{L\times L}$ and $\bm{W}=\bm{w} \bm{w}^H \in \mathbb{C}^{L\times L}$, one can obtain $||\bm{h}_\textit{BS, n}^H \bm{w}||^2={\rm tr}(\bm{H}_n \bm{W})$. Similarly, let $\bm{H}_\textit{eve}=\bm{h}_\textit{BS, eve}^H\bm{h}_\textit{BS, eve}\in \mathbb{C}^{L\times L}$. One has $||\bm{h}_\textit{BS, eve}^H \bm{w}||^2={\rm tr}(\bm{H}_\textit{eve} \bm{W})$. \eqref{eq:beamforming constraint} can be rewritten as
\begin{equation}
    \begin{aligned}
        {\rm tr}(\bm{W}) \le p_\textit{BS}^{tr}.
    \end{aligned}
    \label{transformed beamforming constraint}
\end{equation}
The objective function of \eqref{transformed subprob3} can be transformed into
\begin{equation}\label{SDP objective}
{
    \begin{aligned}
        \tilde{r}^{dl}-r_\textit{eve}^{dl}=\tilde{r}^{dl}-{\rm log}_2(1+\dfrac{{\rm tr}(\bm{H}_\textit{eve} \bm{W})}{\sum_{n'\in \mathcal{N}}(1-\alpha_{n'}){h}_\textit{n', eve}^2 {p_{n'}^{tr}}+\sigma^2}).
    \end{aligned}
    }
\end{equation}
However, \eqref{SDP objective} is non-convex with respect to $\bm{W}$. One can observe that $-{\rm log}_2(1+\frac{{\rm tr}(\bm{H}_\textit{eve} \bm{W})}{\sum_{n'\in \mathcal{N}}(1-\alpha_{n'}){h}_\textit{n', eve}^2 {p_{n'}^{tr}}+\sigma^2})$ is convex with respect to ${\rm tr}(\bm{H}_\textit{eve} \bm{W})$. The lower bound of $-r_\textit{eve}^{dl}$ can be obtained via employing the first order Taylor expansion, which can be formulated as
\begin{equation}
    \begin{aligned}
        &\{-r_\textit{eve}^{dl}\}^{lb}\triangleq-{\rm log}_2(1+\frac{{\rm tr}(\bm{H}_\textit{eve} \bm{W}^{r-1})}{\sum_{n'\in \mathcal{N}}(1-\alpha_{n'}){h}_\textit{n', eve}^2 {p_{n'}^{tr}}+\sigma^2})\\
        &-\frac{\bm{H}_\textit{eve}(\bm{W}-\bm{W}^{r-1})}{{\rm ln}2(\sum_{n'\in \mathcal{N}}(1-\alpha_{n'}){h}_\textit{n', eve}^2 {p_{n'}^{tr}}+\sigma^2+{\rm tr}(\bm{H}_\textit{eve} \bm{W}^{r-1}))},
    \end{aligned}
\end{equation}
where $\bm{W}^{r-1}$ represents the obtained feasible solution in the $(r-1)$-th iteration. According to \eqref{eq:An downlink data rate}, \eqref{beamforming introduced constraint} can be transformed into
\begin{equation}\label{transformed beamforming introduced constraint}
{\small
    \begin{aligned}
        {\rm tr}(\bm{H}_n \bm{W}) \ge (\sum_{n'\in \mathcal{N}}(1-\alpha_{n'}){h}_\textit{n', n}^2 {p_{n'}^{tr}}+\sigma^2)(2^{\tilde{r}^{dl}}-1), n \in \mathcal{N},
    \end{aligned}
    }
\end{equation}
which is a convex constraint. Note that $\bm{W}$ should satisfy
\begin{equation}
\label{rank constraint}
    \begin{aligned}
        {\rm Rank}(\bm{W})=1.
    \end{aligned}
\end{equation}
As such, \eqref{transformed subprob3} can be transformed into
\begin{equation}\label{transformed transformed subprob3}
    \begin{aligned}
        &\underset{\bm{W},\tilde{r}^{dl}}{\max}\; \tilde{r}^{dl}+\{-r_\textit{eve}^{dl}\}^{lb},\\
        &{\rm s.\;t.}\;\; \eqref{transformed beamforming constraint}, \eqref{transformed beamforming introduced constraint},\eqref{rank constraint}.\\
    \end{aligned}
\end{equation}
One can observe that by relaxing the Rank-1 constraint \eqref{rank constraint}, \eqref{transformed transformed subprob3} can be transformed into a convex optimization problem, which can be solved efficiently. Let $\bm{W}^*$ be the obtained solution of \eqref{transformed transformed subprob3}. The optimized beamforming vector $\bm{w}^{r}$ can be obtained via Gaussian randomization based on $\bm{W}^*$ \cite{refnew19}.

\subsubsection{Transmission power optimization}
For any given feasible $\bm{\alpha}$ and $\bm{w}$, \eqref{prob} can be reduced as
\begin{equation} \label{subprob2}
		\begin{aligned}
			&\underset{\bm{p}}{\min}\;\sum_{n \in \mathcal{N}}E_n^\textit{tr},\\
			&{\rm s.\;t.}\; \eqref{eq:latency constraint},\eqref{constraint: transmission power}.\\
		\end{aligned}
\end{equation}
Problem \eqref{subprob2} is difficult to tackle due to the non-smooth objective function and non-convex constraint \eqref{eq:latency constraint}. Two auxiliary variables, i.e., $\dot{c}^{dl}$ and $\dot{\bm{c}}^{ul} \triangleq \{\dot{c}_n^{ul}, n \in \mathcal{N}\}$ are introduced. \eqref{eq:latency constraint} can be transformed into
\begin{equation}
    \begin{aligned}
        \frac{s+D}{\dot{c}^{dl}}+\underset{\forall n \in \mathcal{N}, \alpha_n=1}{\max}\{t_n^\textit{inference}\}+\underset{\forall n \in \mathcal{N}, \alpha_n=1}{\max}\{\frac{o_n}{\dot{c}_n^{ul}}\} \le \tau.
    \end{aligned}
    \label{transformed latency constrant}
\end{equation}

Problem \eqref{subprob2} can be reformulated as
\begin{subequations} \label{transformed subprob2}
    \begin{align}
        &\underset{\bm{p},\dot{c}^{dl},\dot{\bm{c}}^{ul}}{\min}\; \sum_{n \in \mathcal{N}}p_n^{tr}( \alpha_n \frac{o_n}{\dot{c}_n^{ul}} + (1-\alpha_n)(\frac{s+D}{\dot{c}^{dl}}+\max_{\substack{\forall n \in \mathcal{N} \\ \alpha_n=1}}\{\frac{o_n}{\dot{c}_n^{ul}}\})),\\
        &{\rm s.\;t.}\;\;\;\;\;\;\;\;\;\;\;  \eqref{constraint: transmission power},\eqref{transformed latency constrant}\notag\\
        &\;\;\;\;\;\;\;\;\;\;\;\;\;\;\;\;\;\dot{c}^{dl} \ge 0,\label{downlink introduced constraint1}\\
        &\;\;\;\;\;\;\;\;\;\;\;\;\;\;\;\;\;\dot{c}^{dl} \le r^{dl}-r_\textit{eve}^{dl},\label{downlink introduced constraint2}\\
        & \;\;\;\;\;\;\;\;\;\;\;\;\;\;\;\;\;\dot{c}_n^{ul} \ge 0, n \in \mathcal{N},\label{uplink introduced constraint1}\\
        & \;\;\;\;\;\;\;\;\;\;\;\;\;\;\;\;\;\dot{c}_n^{ul} \le r_n^{ul}-r_\textit{n,eve}^{ul}, n \in \mathcal{N}.\label{uplink introduced constraint2}
    \end{align}
\end{subequations}
One can observe that \eqref{downlink introduced constraint2} and \eqref{uplink introduced constraint2} are non-convex constraints. First, we consider the convex approximation of \eqref{downlink introduced constraint2}. According to \eqref{eq:An downlink data rate}-\eqref{eq:achievable rate}, $r^{dl}$ can be rewritten as
\begin{equation}
{\small
    \begin{aligned}
        &r^{dl}=\underset{\forall n \in \mathcal{N}, \alpha_n=1}{\min}\{{\rm log}_2(1+\dfrac{||\bm{h}_\textit{BS, n}^H \bm{w}||^2}{\sum_{n'\in \mathcal{N}}(1-\alpha_{n'}){h}_\textit{n', n}^2 {p_{n'}^{tr}}+\sigma^2})\}\\
        &=\underset{\forall n \in \mathcal{N}, \alpha_n=1}{\min}\{ {\rm log}_2(||\bm{h}_\textit{BS, n}^H \bm{w}||^2+\sum_{n'\in \mathcal{N}}(1-\alpha_{n'}){h}_\textit{n', n}^2 {p_{n'}^{tr}}+\sigma^2)\\
        &-  {\rm log}_2(\sum_{n'\in \mathcal{N}}(1-\alpha_{n'}){h}_\textit{n', n}^2 {p_{n'}^{tr}}+\sigma^2)    \}.
    \end{aligned}
    }
\end{equation}
One can observe that ${\rm log}_2(||\bm{h}_\textit{BS, n}^H \bm{w}||^2+\sum_{n'\in \mathcal{N}}(1-\alpha_{n'}){h}_\textit{n', n}^2 {p_{n'}^{tr}}+\sigma^2)$ and ${\rm log}_2(\sum_{n'\in \mathcal{N}}(1-\alpha_{n'}){h}_\textit{n', n}^2 {p_{n'}^{tr}}+\sigma^2)$ are convex with respect to $\bm{p}$. The lower bound of $r^{dl}$  can be obtained via employing the first Taylor expansion of ${\rm log}_2(||\bm{h}_\textit{BS, n}^H \bm{w}||^2+\sum_{n'\in \mathcal{N}}(1-\alpha_{n'}){h}_\textit{n', n}^2 {p_{n'}^{tr}}+\sigma^2)$, which can be given as \eqref{lower bound of r^dl},
\begin{figure*}[!t]
    \begin{equation} \label{lower bound of r^dl}
    {\small
        \begin{aligned}
           r^{dl,lb}&=\underset{\substack{\forall n \in \mathcal{N} \\ \alpha_n=1}}{\min}\{ {\rm log}_2(||\bm{h}_\textit{BS, n}^H \bm{w}||^2+\sum_{n'\in \mathcal{N}}(1-\alpha_{n'}){h}_\textit{n', n}^2 {p_{n'}^{tr,r-1}}+\sigma^2)+\frac{\sum_{n'\in \mathcal{N}}(1-\alpha_{n'}){h}_\textit{n', n}^2(p_{n'}^{tr}-p_{n'}^{tr,r-1})}{{\rm ln}2(||\bm{h}_\textit{BS, n}^H \bm{w}||^2+\sum_{n'\in \mathcal{N}}(1-\alpha_{n'}){h}_\textit{n', n}^2 {p_{n'}^{tr,r-1}}+\sigma^2)}\\
        &-  {\rm log}_2(\sum_{n'\in \mathcal{N}}(1-\alpha_{n'}){h}_\textit{n', n}^2 {p_{n'}^{tr}}+\sigma^2)    \},
        \end{aligned}}
    \end{equation}
    \hrulefill
\end{figure*}
where $\bm{p}^{r-1}\triangleq\{p_{n}^{tr,r-1}, n \in \mathcal{N}\}$ represents the obtained feasible solution in the $r-1$-th iteration. Similarly, $r_\textit{eve}^{dl}$ can be rewritten as ${\rm log}_2(\sum_{n'\in \mathcal{N}}(1-\alpha_{n'}){h}_\textit{n', eve}^2 {p_{n'}^{tr}}+\sigma^2+||\bm{h}_\textit{BS, eve}^H \bm{w}||^2)-{\rm log}_2(\sum_{n'\in \mathcal{N}}(1-\alpha_{n'}){h}_\textit{n', eve}^2 {p_{n'}^{tr}}+\sigma^2)$ and the upper bound of $r_\textit{eve}^{dl}$, denoted by $r_\textit{eve}^{dl,ub}$ can be obtained by employing the first Taylor expansion of ${\rm log}_2(\sum_{n'\in \mathcal{N}}(1-\alpha_{n'}){h}_\textit{n', eve}^2 {p_{n'}^{tr}}+\sigma^2)$. \eqref{downlink introduced constraint2} can be transformed into 
\begin{equation}\label{ transformed downlink introduced constraint2}
    \begin{aligned}
        \dot{c}^{dl} \le r^{dl,lb}-r_\textit{eve}^{dl,ub},
    \end{aligned}
\end{equation}
which is a convex constraint. Then, the convex approximation of \eqref{uplink introduced constraint2} is considered. The lower bound and upper bound of $r_n^{ul}$ and $r_\textit{n,eve}^{ul}$, denoted by $r_n^{ul,lb}$ and $r_\textit{n,eve}^\textit{ul,ub}$ respectively, can be obtain in the similar fashion to $r^{dl,lb}$ and $r_\textit{eve}^{dl,ub}$. As such, \eqref{uplink introduced constraint2} can be transformed into
\begin{equation}\label{ transformed uplink introduced constraint2}
    \begin{aligned}
        \dot{c}_n^{ul} \le r_n^{ul,lb}-r_n^\textit{eve,ub}.
    \end{aligned}
\end{equation}
Consequently, \eqref{transformed subprob2} can be transformed into
\begin{equation} \label{transformed transformed subprob2}
    \begin{aligned}
        &\underset{\bm{p},\dot{c}^{dl},\dot{\bm{c}}^{ul}}{\min}\; \sum_{n \in \mathcal{N}}p_n^{tr}( \alpha_n \frac{o_n}{\dot{c}_n^{ul}} + (1-\alpha_n)(\frac{s+D}{\dot{c}^{dl}}+\max_{\substack{\forall n \in \mathcal{N} \\ \alpha_n=1}}\{\frac{o_n}{\dot{c}_n^{ul}}\})),\\
        &{\rm s.\;t.}\;\;\;\;\;\;\;\;\;\;\;\;\;\;\;  \eqref{constraint: transmission power},\eqref{transformed latency constrant},\eqref{downlink introduced constraint1},\eqref{uplink introduced constraint1},\eqref{ transformed downlink introduced constraint2},\eqref{ transformed uplink introduced constraint2}.
    \end{aligned}
\end{equation}
Problem \eqref{transformed transformed subprob2} is a convex optimization problem, which can be solved efficiently. The optimized transmission power $\bm{p}^{r}$ for the current iteration can be obtained.

\subsubsection{The overall process of ASC}
The detailed information of ASC is summarized in \textbf{Algorithm \ref{the proposed solution}}. Let $r^\textit{max}$ be the maximum number of iterations of the proposed solution. The feasible solution of \eqref{prob} can be obtained when the proposed solution reaches convergence with $r=r^\textit{max}$. The complexity of the proposed ADMM-based algorithm can be estimated as $\mathcal{O}(r_\textit{ADMM}^\textit{max}(r_\textit{LR}^\textit{max}N+N^{3.5}))$. The complexities of solving \eqref{transformed transformed subprob3} and \eqref{transformed transformed subprob2} can be given as $\mathcal{O}(L^{3.5})$ and $\mathcal{O}(N^{3.5})$, respectively. As such, the complexity of ASC can be roughly given as $\mathcal{O}(r^\textit{max}(r_\textit{ADMM}^\textit{max}(r_\textit{LR}^\textit{max}N+N^{3.5})+L^{3.5}+N^{3.5}))$.

\begin{algorithm}
		\caption{The framework of ASC}
		\label{the proposed solution}
		Initialize $\bm{\alpha}^{0}$, $\bm{w}^0$, $\bm{p}^0$ and $r=0$.\\
        \While{$r \le r^\textit{max}$}{
		Fix $\bm{w}^{r-1}$ and $\bm{p}^{r-1}$, utilize \textbf{Algorithm} \ref{ADMM}  to obtain $\bm{\alpha}^{r}$.\\
        Fix $\bm{\alpha}^{r}$ and $\bm{p}^{r-1}$, solve \eqref{transformed transformed subprob3} to obtain $\bm{w}^r$.\\
        Fix $\bm{\alpha}^r$ and $\bm{w}^r$, solve \eqref{transformed transformed subprob2} to obtain $\bm{p}^r$.\\
        $r \leftarrow r+1$.\\
		}
        \textbf{Return} $\bm{\alpha}^{r^\textit{max}}$, $\bm{w}^{r^\textit{max}}$ and $\bm{p}^{r^\textit{max}}$.\\
	\end{algorithm}

\begin{figure}[!h]
		\centering
		\includegraphics[width=1.02\linewidth]{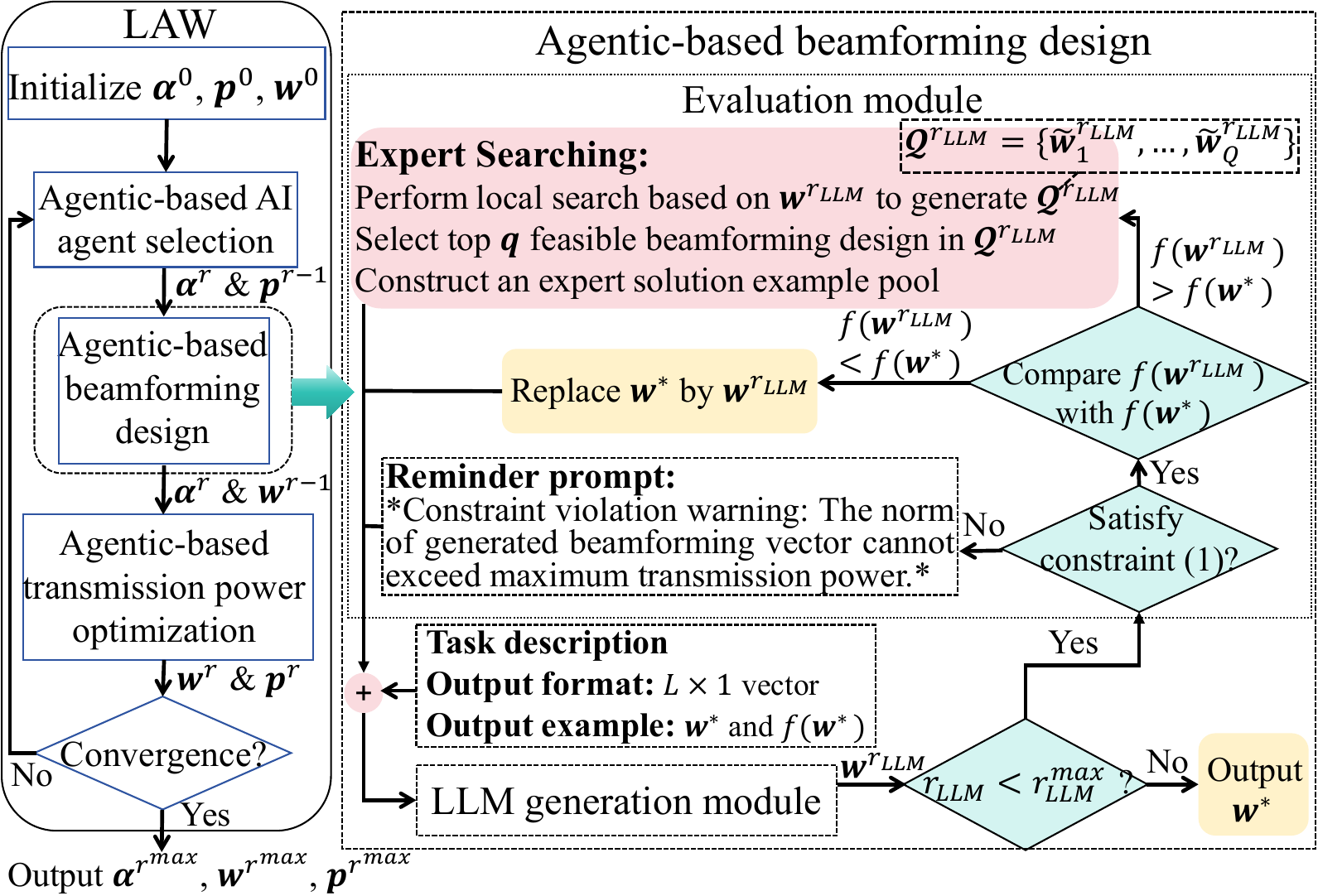}
		\caption{The framework of LAW.}
        \label{fig:LAW}
\end{figure}
\subsection{LAW}
In this subsection, we introduce LAW to tackle \eqref{prob}. The framework of LAW is demonstrated in Fig. \ref{fig:LAW}. Similar to the introduced ASC, LAW decouples \eqref{prob} into three sub-problems and each sub-problem is solved by the formulated LLM optimizer based on an adaptive reflection agentic workflow in an iterative manner.

In particular, the LLM optimizer includes two parts, i.e., LLM generation module and evaluation module. The LLM generation module is responsible for generating feasible solutions to three sub-problems. Consider the beamforming design sub-problem shown in Fig. \ref{fig:LAW}. The input prompt of LLM generation module consists of task description, output formatting constraints, and output example. In the initialization stage, task description outlining the objective function, constraints and key parameters, output format of \eqref{subprob3}, and the output example are input to LLM generation module. The output example is denoted by $\bm{w}^*$ and $f(\bm{w}^*)$, which can enhance the continuity in each interaction with LLM \cite{ref26}. One should notice that $f(\cdot)$ is the objective function of \eqref{prob} and $f(\bm{w}^*)$ can be obtained via fixing AI agent selection and transmission power.

The evaluation module is responsible for evaluating the quality of LLM generated feasible solution and providing reflection to dynamically refine the LLM's output and enhance the solution quality through closed-loop feedback. Denote the feasible beamforming vector generated by LLM generation module in the $r_\textit{LLM}$-th iteration by $\bm{w}^{r_\textit{LLM}}$. First, the evaluation module analyzes whether $\bm{w}^{r_\textit{LLM}}$ satisfies \eqref{eq:beamforming constraint}. If \eqref{eq:beamforming constraint} is violated, the evaluation module generates a reminder prompt to emphasize that the LLM-generated solution must adhere to the constraint, and sets $f(\bm{w}^{r_\textit{LLM}})=+\infty$. Next, consider two conditions, if $f(\bm{w}^{r_\textit{LLM}})< f(\bm{w}^*)$, One can set $\bm{w}^*=\bm{w}^{r_\textit{LLM}}$ and add updated $\bm{w}^*$ to the input prompts in the next iteration. If $f(\bm{w}^{r_\textit{LLM}}) \ge f(\bm{w}^*)$, an expert searching mechanism is triggered to further exploit the solution space. Specifically, a local neighborhood search is performed based on $\bm{w}^{r_\textit{LLM}}$ to generate a candidate pool $\bm{Q}^{r_\textit{LLM}}=\{\tilde{\bm{w}}_1^{r_\textit{LLM}},\cdots \tilde{\bm{w}}_Q^{r_\textit{LLM}}\}$. Each candidate is analyzed by their fitness values based on $f(\cdot)$, and the top $q$ feasible solutions in the candidate pool are selected to construct an expert solution example pool. By incorporating these high-quality solutions into the next iteration's prompt, the framework significantly enhances the search capability of the LLM. The agentic-based beamforming design can be regarded as convergence when $r_\textit{LLM}=r_\textit{LLM}^\textit{max}$, where $r_\textit{LLM}^\textit{max}$ represents the maximum number of iterations.

The overall frameworks of agentic-based AI agent selection and transmission power optimization are identical to agentic-based beamforming design. The proposed LAW successively optimize AI agent selection, beamforming design and transmission power until $r=r^\textit{max}$, where $r^\textit{max}$ denotes the maximum number of iterations of the proposed LAW. As such, the feasible solution of \eqref{prob} can be obtained. One should notice that LAW can be employed by the supervisor AI agent directly. By utilizing its equipped LLM as an intelligent optimizer, the supervisor AI agent can dynamically allocate communication and computing resources to other AI agents.

\section{Performance Evaluation}\label{performance evaluation}
In this section, we investigate the effectiveness of the proposed solutions for the proposed secure wireless agentic AI network by comparing them with several selected benchmark algorithms. All experiments are conducted in Python 3.13.9 with CVXPY toolbox on a PC with Intel Core i7-12700K and 16GB RAM. The proposed LAW employs GPT-4-turbo as an optimizer via commercial API in the performance analysis. The key simulation parameters are summarized as follows. The coordinates of BS and eve are set as $[0,0]$ m and $[150,150]$ m, respectively. All AI agents are randomly distributed in a square area of 100 m $\times$ 100 m. We assume that wireless channels in the secure wireless agentic AI network follow Rayleigh fading with a mean of $10^{-\frac{\textit{PL}(d)}{20}}$, where $\textit{PL}(d)=32.4+20{\rm lg}(f_\textit{carrier})+20 {\rm lg}(d)$. $d$ is the distance between the transmitter and the receiver. $f_\textit{carrier}$ denotes the carrier frequency and is set to 3.5 GHz. The lengths of input prompt and reasoning output of a reasoning task are set as 2048 and 4096 tokens, respectively. Each token is represented by a 16-bit integer index. Other important parameters are given in TABLE \ref{tab:simulation_parameters}.
	
\begin{table}[htbp]
  \centering
  \small
  \caption{Simulation parameters \cite{refnew18,ref28}.}
  \label{tab:simulation_parameters}

  \begin{tabular*}{\columnwidth}{
    @{\hspace{3\tabcolsep}\extracolsep{\fill}} 
    cc
    !{\hspace{\tabcolsep}\vrule\hspace{\tabcolsep}} 
    cc
    @{\hspace{\tabcolsep}} 
  }
    \toprule
    \textbf{Notation} & \textbf{Value} & \textbf{Notation} & \textbf{Value} \\
    \midrule
    $p_{\text{BS}}^{\text{tr}}$ & 10 W       & $B$        & 5 MHz \\
    $p_n^{\text{tr,max}}$       & 2 W        & $\sigma^2$  & -107 dBm \\
    $p_n^{\text{exe}}$          & [10, 20] W & $k_n$       & [2560, 4096] \\
    $D$                         & 10 Mbits   & $\tau$      & 3000 s \\
    \bottomrule
  \end{tabular*}
\end{table}

\subsection{Analysis of execution energy consumption}
We first evaluate the execution energy consumption performance by comparing the proposed ASC and LAW against two benchmark schemes: the genetic algorithm (GA) and Random AI agent selection (RandAS). These benchmarks utilize GA and random selection for AI agent assignment, respectively, while the optimization of BS beamforming and AI agent transmission powers is identical to ASC.
\begin{figure}[!h]
		\centering
		\includegraphics[width=0.85\linewidth]{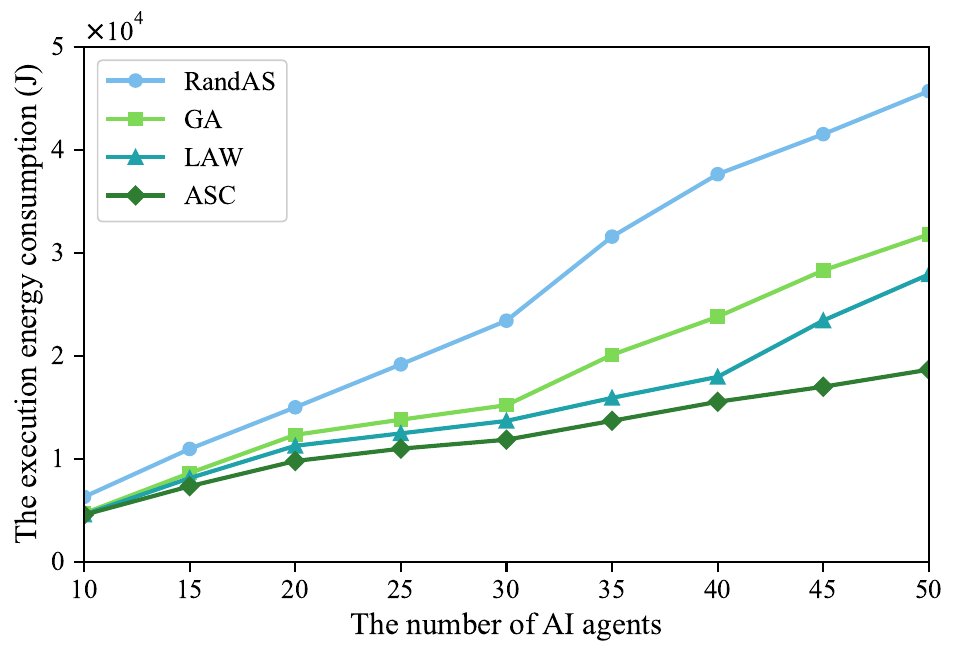}
		\caption{The execution energy consumption versus different numbers of AI agents.}
		\label{fig:EexeVSnum}
\end{figure}

Fig. \ref{fig:EexeVSnum} demonstrates the execution energy consumption versus the number of AI agents. One can observe that the execution energy consumption increases as the number of AI agents increases. In particular, ASC achieves the lowest execution energy consumption among the four schemes with approximately 1.19 $\times 10^4$ J and $1.87 \times 10^4$ J when $N=30$ and 50, respectively. LAW realizes slightly higher execution energy consumption in comparison to ASC with corresponding values of 1.37 $\times 10^4$ J and 2.79$\times 10^4$ J. RandAS realizes the worst performance with 2.34$\times 10^4$ J and 4.57$\times 10^4$ J when $N=30$ and 50. This is because ASC and LAW can dynamically select AI
agents to participate in cooperative reasoning according to their reasoning accuracy and computation resources. In this way, the execution energy consumption can be reduced significantly while satisfying the reasoning accuracy constraint.

\begin{table}[htbp]
  \centering
  \small
  \caption{The execution energy consumption (J) under different minimum numbers of reasoning AI agents ($N_{\text{min}}$).}
  \label{tab:energy_consumption}
  \begin{tabular*}{\columnwidth}{@{\extracolsep{\fill}}lccc}
    \toprule
    Schemes & $N_{\text{min}}=10$ & $N_{\text{min}}=15$ & $N_{\text{min}}=20$ \\
    \midrule
    RandAS  & 23417 & 41107 & 48999 \\
    GA      & 15019 & 29983 & 38402 \\
    LAW   & 13693 & 28436 & 36113 \\
     ASC   & 11868 & 26673 & 34996 \\
    \bottomrule
  \end{tabular*}
\end{table}

\begin{table}[htbp]
  \centering
  \small
  \caption{The execution energy consumption (J) under different average reasoning accuracy thresholds ($Acc_{\text{min}}$).}
  \label{tab:energy_accuracy}
  \begin{tabular*}{\columnwidth}{@{\extracolsep{\fill}}lccc}
    \toprule
    Schemes & $Acc_{\text{min}}=60\%$ & $Acc_{\text{min}}=70\%$ & $Acc_{\text{min}}=80\%$ \\
    \midrule
    RandAS & 21222 & 23417 & 27520 \\
    GA     & 13120 & 15219 & 16189 \\
    LAW  & 12022 & 14242 & 14959 \\
    ASC  & 10423 & 11868 & 12966 \\
    \bottomrule
  \end{tabular*}
\end{table}

The typical number of AI agents, i.e., $N=30$ is employed for further analysis of execution energy consumption. Since the minimum number of reasoning AI agents plays a pivotal role in guaranteeing the reliability and robustness of cooperative reasoning, we investigate the impact of varying the minimum number of participating AI agents on the performance of ASC and LAW. TABLE \ref{tab:energy_consumption} shows the execution energy consumption versus different minimum numbers of reasoning AI agents. ASC realizes the best performance compared to other three schemes. Specifically, ASC achieves $1.19 \times 10^4$ J and $3.50 \times 10^4$ J when $N_\textit{min}=10$ and 20, respectively, followed by LAW and GA with corresponding values of $1.37\times 10^4$ J and $3.61\times 10^4$ J, $1.50\times 10^4$ J and $3.84\times 10^4$ J. This phenomenon can be explained by the fact that the computational energy consumption for LLM inference is typically larger than the communication energy overhead. As the number of AI agents involved in cooperative reasoning increases, the total energy consumption escalates significantly. Therefore, the number of reasoning AI agents should be reasonably determined.

Furthermore, we evaluate the impact of the required average reasoning accuracy on the performance of the proposed solutions. The typical value $N_\textit{min}=10$ is adopted in the following study. TABLE \ref{tab:energy_accuracy} shows the execution energy consumption under different average reasoning accuracy threshold. It can be seen that a higher average reasoning accuracy threshold leads to an increase in execution energy consumption. In particular, ASC realizes $1.04 \times 10^4$ J and $1.30 \times 10^4$ J when $\textit{Acc}_\textit{min}=60\%$ and $80\%$, respectively. LAW and GA achieve slightly higher execution energy consumption in comparison to ASC, with corresponding values of $1.20 \times 10^4$ J and $1.49 \times 10^4$ J, $1.31 \times 10^4$ J and $1.62 \times 10^4$ J. RandAS realizes the highest execution energy consumption among four schemes with $2.12\times 10^4$ J and $2.75 \times 10^4$ J when $\textit{Acc}_\textit{min}=60\%$ and $80\%$, respectively. This is because $A_s$ prioritizes  to select AI agents equipped with larger-scale LLMs to satisfy the stricter accuracy threshold, thereby increasing the network's computational cost and execution energy consumption.

\subsection{Analysis of transmission energy consumption}
In this subsection, we study the transmission energy consumption performance of the proposed solutions in comparison to three benchmark schemes, i.e., differential evolution scheme (DE), fixed transmission power scheme (FixedTP) and fixed beamforming scheme (FixedBF). In particular, DE shares the identical AI agent selection scheme with ASC and utilizes DE algorithm to jointly optimize BS beamforming and AI agent transmission powers. FixedTP sets the transmission power of each AI agent to $60\%$ of the maximum transmission power and the optimization of AI agent selection and BS beamforming is identical to ASC. FixedBF allocates BS transmission power equally to each antenna and adopts the same method as ASC for AI agent selection and transmission power optimization.

\begin{figure}[!h]
		\centering
		\includegraphics[width=0.85\linewidth]{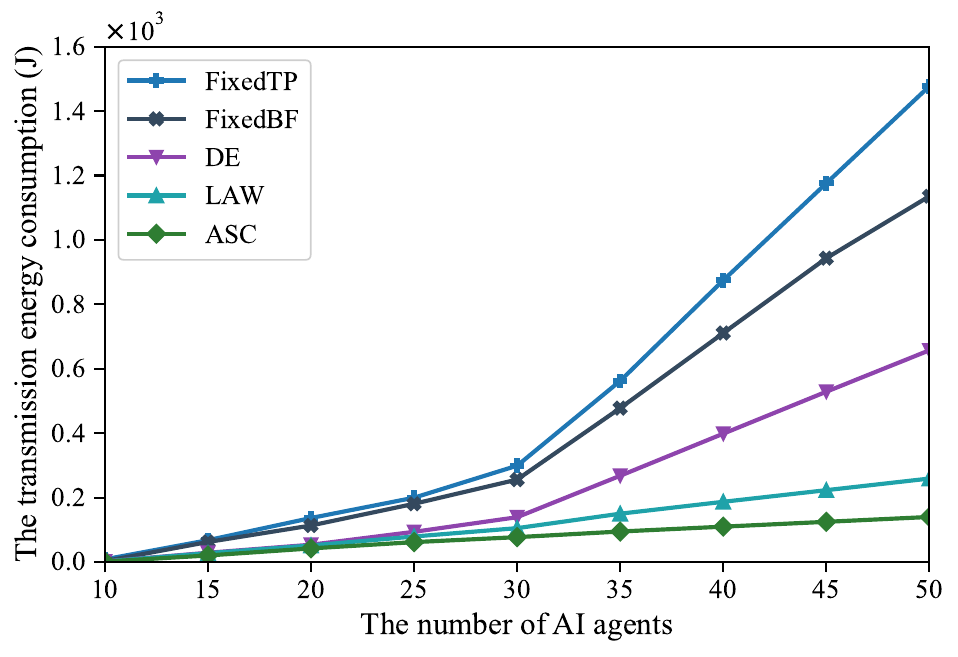}
		\caption{The transmission energy consumption versus different numbers of AI agents.}
		\label{fig:EtrVSnum}
\end{figure}

Fig. \ref{fig:EtrVSnum} plots the transmission energy consumption under different numbers of AI agents. ASC realizes the best performance among five schemes with around 77 and 140 J when $N=30$ and 50, respectively. LAW achieves slightly higher transmission energy consumption with corresponding values of 105 J and 260 J. FixedTP exhibits 199 J and 1477 J when $N=30$ and 50, respectively, which is the highest among five schemes. This is because ASC and LAW can dynamically optimize AI agent transmission powers according to AI agent selection and channel states. As such, these solutions can achieve the required uplink secrecy capacity with minimal power overhead, thereby reducing the overall energy consumption while satisfying the latency constraint.

\begin{table}[htbp]
  \centering
  \small
  \caption{The transmission energy consumption (J) under different numbers of BS antennas ($L$).}
  \label{tab:trans_energy_L}
  \begin{tabular*}{\columnwidth}{@{\extracolsep{\fill}}lccc}
    \toprule
    Schemes & $L=10$ & $L=20$ & $L=30$ \\
    \midrule
    FixedTP & 310.76 & 299.20 & 225.38 \\
    FixedBF & 260.27 & 246.01 & 171.85 \\
    DE      & 155.95 & 139.21 & 54.967  \\
    LAW   & 146.01 & 105.21 & 53.785  \\
    ASC   & 109.87 & 77.219  & 49.872  \\
    \bottomrule
  \end{tabular*}
\end{table}

A baseline configuration of $N=30$ is adopted for the evaluation of the impact of different numbers of BS antennas on the transmission energy consumption. TABLE \ref{tab:trans_energy_L} presents the impact of the number of BS antennas on transmission energy consumption. One can observe that the transmission energy consumption decreases as the number of BS antennas increases. In particular, ASC and LAW achieve 110 J and 50 J, 146 J and 54 J when $L=10$ and 30, respectively, followed by DE with corresponding values of 156 J and 55 J. FixedTP realizes the worst performance with 311 J and 225 J when $L=10$ and 30, respectively. One should notice that the proposed solution can dynamically optimize BS beamforming to achieve better downlink secrecy capacity. The energy consumption of AI agents functioning as friendly jammers in the downlink transmission process can be reduced significantly. Moreover, with more antennas, BS can focus the transmission beam more precisely on the selected AI agents, which may reduce signal leakage to the eavesdropper and improve the downlink secrecy capacity.

\subsection{Analysis of reasoning performance on Public benchmarks}
To evaluate the practical reasoning capabilities of the proposed secure agentic AI network with ASC and LAW in real-world scenarios, we construct an agentic AI system comprising 30 AI agents, where two sets of 12 AI agents are equipped with Qwen-0.6B and Qwen-1.7B models. One set of 6 AI agents is equipped with Qwen-4B models. Each AI agent is deployed on an NVIDIA Tesla P100 GPU with 16G VRAM. The number of AI agents involved in cooperative reasoning is set to 10. The reasoning accuracy on three widely used datasets: ARC-E, ARC-C, and BoolQ, which represent different levels of reasoning difficulty, is investigated \cite{refnew18}.
\begin{table}[htbp]
  \centering
  \small
  \caption{The reasoning accuracy under different public benchmarks.}
  \label{tab:Acc}
  \begin{tabular*}{\columnwidth}{@{\extracolsep{\fill}}lccc}
    \toprule
    Schemes & ARC-E & ARC-C & BoolQ \\
    \midrule
    ASC   & 85\% & 75\% & 83\%  \\
    LAW   & 76\% & 65\%  & 82\%  \\
    GA      & 66\% & 56\% & 80\%    \\
    RandAS  &56\% & 41\% & 75\%     \\
    \bottomrule
  \end{tabular*}
\end{table}

TABLE \ref{tab:Acc} shows the reasoning accuracy of the proposed secure agentic AI network under different public benchmarks. In ARC-E and ARC-C, the proposed ASC and LAW achieve significantly higher accuracy compared to that of RandAS. In particular, ASC and LAW realize 85\% and 75\%, 76\%  and 65\%, respectively. This is because ASC and LAW can dynamically select participating AI agents according to their computational resources and the reasoning accuracy predicted by the scaling law. As such, the proposed schemes can realize satisfactory reasoning accuracy while minimizing energy consumption. Additionally, one should note that the performance gaps between four schemes on BoolQ are not as significant as in the other datasets. This phenomenon may be attributed to the fact that BoolQ questions are binary classification problems. The majority voting mechanism remains robust in improving reasoning accuracy even when hallucinations occur in some participating AI agents. However, for the more challenging ARC-C dataset with a larger solution space, RandAS performs significantly worse than the other schemes because it cannot dynamically select AI agents.

\section{Conclusion}\label{Conclusion}
We propose a novel secure wireless agentic AI network, where one supervisor AI agent and other AI agents are jointly deployed for QoS provisioning of users' reasoning tasks. In particular, the supervisor AI agent dynamically selects multiple AI agents to engage in cooperative reasoning and the unselected AI agents function as friendly jammers to mitigate eavesdropping. To prolong the service duration of AI agents, an energy minimization problem is formulated by jointly considering AI agent selection, BS beamforming and AI agent transmission powers, subject to a list of QoS constraints. To solve the formulated challenging problem, two resource allocation methods, ASC and LAW, are introduced. These methods first decouple the original problem into three sub-problems. ASC optimizes three sub-problems by the proposed ADMM-based algorithm, SDR and SCA in an alternating manner, while LAW solves each sub-problem via the formulated LLM optimizer following an agentic workflow. Experimental results verify that the proposed schemes can significantly reduce AI agents energy consumption while promising satisfactory QoS performance in comparison to other benchmark algorithms.

\bibliographystyle{IEEEtran} 

\bibliography{references}

	\newpage

	\vfill
	
\end{document}